%% file: main.tex
\documentclass[10pt,twocolumn,letterpaper]{article}

\usepackage{cvpr}              
\usepackage{natbib}
\setcitestyle{square}
\setcitestyle{comma}
\usepackage{epigraph}

\input{preamble}

\usepackage{booktabs}
\usepackage[normalem]{ulem}
\useunder{\uline}{\ul}{}

\definecolor{cvprblue}{rgb}{0.21,0.49,0.74}
\usepackage[pagebackref,breaklinks,colorlinks,citecolor=cvprblue]{hyperref}

\def\paperID{*****} 
\def\confName{CVPR}
\def\confYear{2024}

\title{LLM Inference Unveiled: Survey and Roofline Model Insights}

\author{
  Zhihang Yuan$^{1,}$\thanks{Equal contribution. Zhihang Yuan (hahnyuan@gmail.com) is the project leader.}~~~~~~~~~Yuzhang Shang$^{2,*}$~~~~~~~~~Yang Zhou$^{3,*}$~~~~~~~~~Zhen Dong$^{8}$\\
  Zhe Zhou$^{4}$~~~~~~~~~Chenhao Xue$^{4}$~~~~~~~~~Bingzhe Wu$^{5}$~~~~~~~~~ Zhikai Li$^{6}$~~~~~~~~~Qingyi Gu$^{6}$\\
  Yong Jae Lee$^{7}$~~~~~~~~~Yan Yan$^{2}$~~~~~~~~~Beidi Chen$^{3}$~~~~~~~~~ Guangyu Sun$^{4}$~~~~~~~~~Kurt Keutzer$^{8}$ \vspace{8pt} \\
  \fontsize{11}{15}\selectfont \textit{$^{1}$Infinigence-AI , $^{2}$Illinois Institute of Technology, $^{3}$Carnegie Mellon University, $^{4}$Peking University, $^{5}$Tencent AI Lab,}\\
  \fontsize{11}{15}\selectfont \textit{$^{6}$Institute of Automation, CAS, $^{7}$University of Wisconsin, Madison, $^{8}$University of California, Berkeley.}
}

\begin{document}
\maketitle
\input{sections/0_abstract}
\input{sections/1_intro}

\input{sections/2_challenges}
\input{sections/3_compression}
\input{sections/4_decoding}

\input{sections/5_system}


\input{sections/6_hardware}
\input{sections/8_discussion}

\input{sections/9_conclusion}

\clearpage
{
    \small
    \bibliographystyle{apalike}
    \bibliography{main}
}


\end{document}

%% file: preamble.tex
\usepackage[dvipsnames]{xcolor}


%% file: sections/0_abstract.tex
\begin{abstract}
The field of efficient Large Language Model (LLM) inference is rapidly evolving, presenting a unique blend of opportunities and challenges. 
Although the field has expanded and is vibrant, there hasn't been a concise framework that analyzes the various methods of LLM Inference to provide a clear understanding of this domain.
Our survey stands out from traditional literature reviews by not only summarizing the current state of research but also by introducing a framework based on Roofline model for systematic analysis of LLM inference techniques. 
This framework identifies the bottlenecks when deploying LLMs on hardware devices and provides a clear understanding of practical problems, such as why LLMs are memory-bound, how much memory and computation they need, and how to choose the right hardware.
We systematically collate the latest advancements in efficient LLM inference, covering crucial areas such as model compression (e.g., quantization), algorithm improvements (e.g., speculative decoding), and both system and hardware-level enhancements (e.g., operator fusion). 
Our survey stands out by analyzing these methods with Roofline model, helping us understand their impact on memory access and computation.
This distinctive approach not only showcases the current research landscape but also delivers valuable insights for practical implementation, positioning our work as an indispensable resource for researchers new to the field as well as for those seeking to deepen their understanding of efficient LLM deployment. 
The analyze tool, \href{https://github.com/hahnyuan/LLM-Viewer}{LLM-Viewer}, is open-sourced.
\end{abstract}

%% file: sections/1_intro.tex
\section{Introduction}

\begin{figure}[t]
    \centering
    \includegraphics[width=0.95\linewidth]{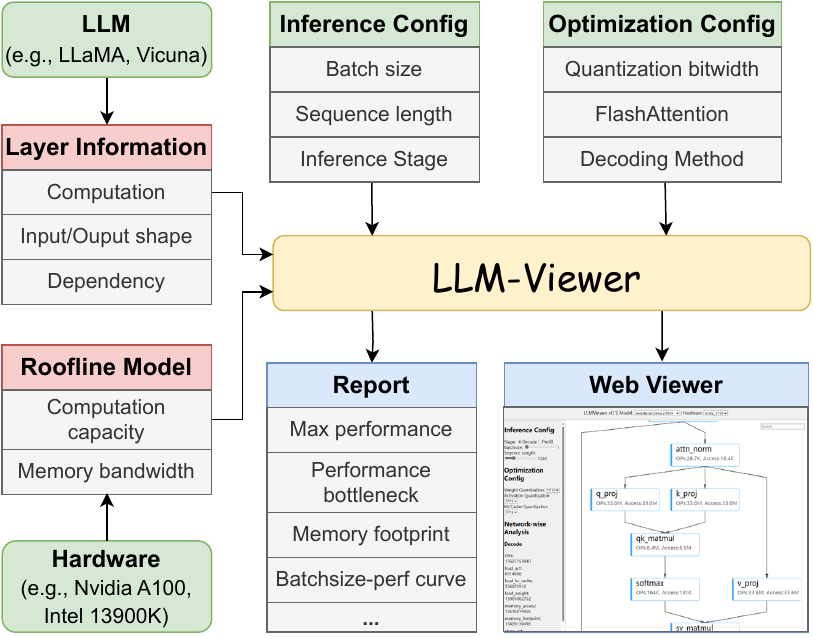}
    \caption{\textbf{Workflow of our designed LLM-Viewer.} Input:  the configuration details for the intended LLM deployment and specific hardware device information. Upon receiving these inputs, the LLM-Viewer is designed to precisely analyze and identify the bottlenecks associated with deploying the given LLM on the specified hardware device, facilitating targeted optimizations for efficient LLM inference.}
    \vspace{-0.2in}
    \label{fig:llmviewer_workflow}
\end{figure}

\begin{figure*}[t]
    \centering
    \includegraphics[width=0.96\textwidth]{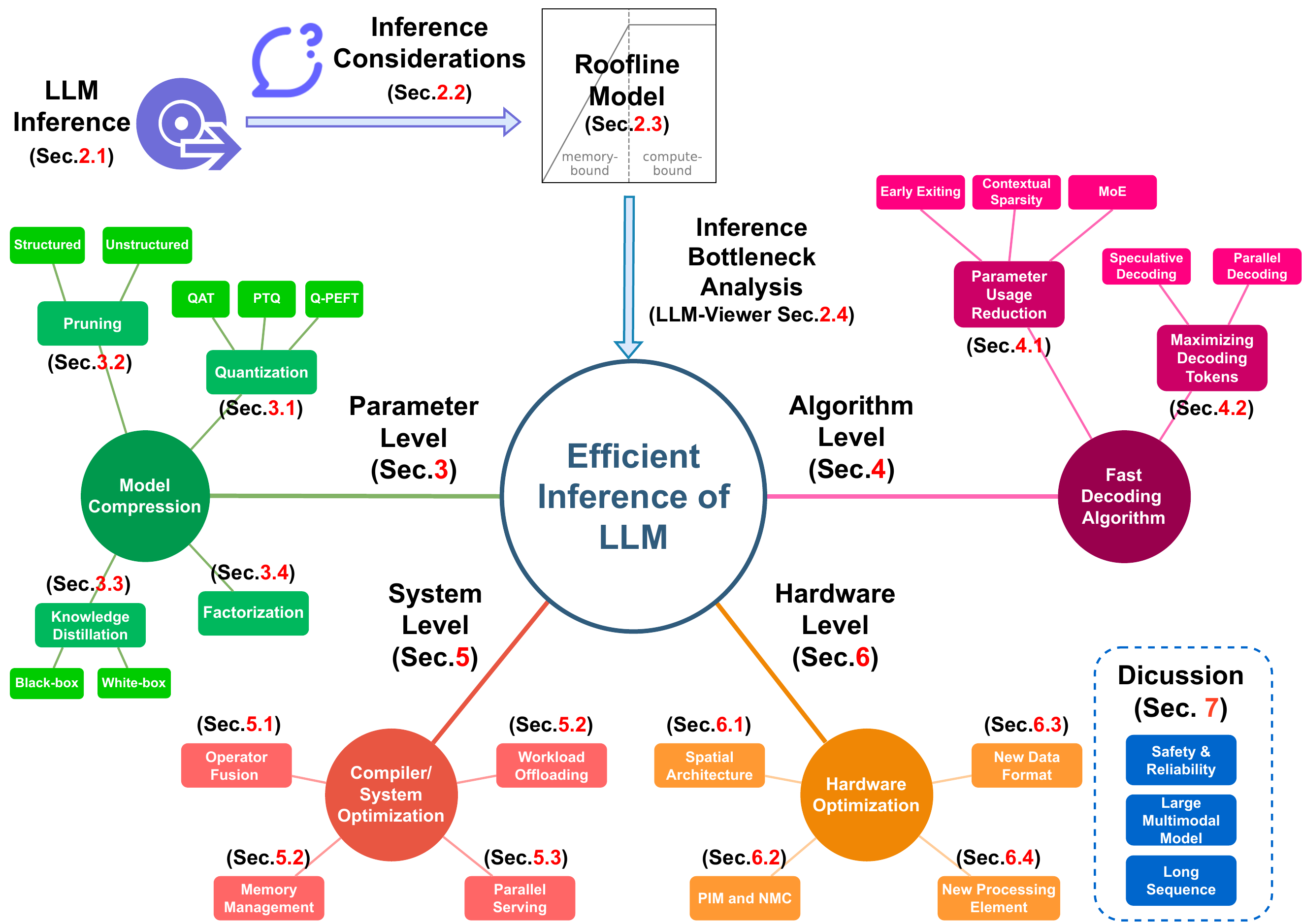}
    \caption{\textbf{Mind-map of the Survey on Efficient LLM Inference}. Our survey diverges from traditional surveys by focusing on the practical aspects of LLM inference. Specifically, we identify and analyze the challenges associated with LLM inference. Subsequently, we introduce a specially developed Roofline model to pinpoint the bottlenecks in LLM inference processes (Sec.\ref{sec:challenges}). The survey categorizes strategies for improving LLM inference efficiency into four main areas: Parameter Reduction (Sec.\ref{sec:compression}), Fast Decoding Algorithm Design (Sec.\ref{sec:fast-decoding}), System-Level Optimization (Sec.\ref{sec:system_optimization}), and Hardware-Level Optimization (Sec.\ref{sec:hardware-optimization}), providing a comprehensive framework for addressing the complexities of efficient LLM deployment.}
    \vspace{-0.2in}
    \label{fig:mindmap}
\end{figure*}

Large Language Models (LLMs) have become a cornerstone of AI advancement in recent years, reshaping the landscape of machine learning and natural language processing (NLP)~\citep{zhao2023survey}. This trend can be traced to the success of revolutionary models like ChatGPT~\citep{brown2020language,ouyang2022training}, which produce very human-like text through their exceptional understanding and generation abilities. 
Following ChatGPT, other notable LLMs such as OPT~\citep{zhang2022opt}, BLOOM~\citep{scao2022bloom}, and Llama~\citep{touvron2023llama1,touvron2023llama2} have emerged, further solidifying the consensus that larger models often lead to enhanced capabilities. Therefore, models with tens of billions of parameters are becoming increasingly common. 
As a result of the vast size of these models, they present considerable inference challenges, not only for devices with limited computational capabilities, but also for the most advanced hardware. 
Because of their complexity and scale, as well as their energy and computational demands, these models are difficult to deploy in real-world situations. 
Additionally, the resource-intensive nature of these models raises concerns about energy consumption, scalability, and accessibility. The situation is particularly challenging for smaller organizations and communities with fewer computing resources than large corporations. Therefore, these challenges emphasize the need for innovative solutions to make LLM inference more universally accessible and sustainable.

Numerous methods have been developed to address the challenges of deploying LLM. The field of efficient LLM inference has grown exponentially in the last two years, presenting both opportunities and challenges. 
While the burgeoning volume of research demonstrates the field's vibrancy, it can inadvertently mask key trends and slow advancements. 
A critical gap in existing literature is the absence of a systematic and practical framework for unified analysis and comprehensive solution development.
To bridge this gap, our work offers a comprehensive overview of the current state of research in efficient LLM inference, with a unique focus on its practice-driven characteristics. Diverging from traditional literature reviews, our work not only discusses existing research but also introduces a specifically developed Roofline model. This model is designed to analyze bottlenecks in LLM deployments, a crucial step we believe is vital for practical application and optimization as shown in Figure~\ref{fig:llmviewer_workflow}.
Our work is the first, to our knowledge, that provides such a tool for analyzing the intricacies of inferring LLMs on hardware devices, systematically collating the latest advancements in efficient LLM inference. We delve deep into deployment challenges, particularly emphasizing inference efficiency. Our discussion spans various areas, including model compression, decoding algorithm refinement, system-level and hardware-level enhancements, as illustrated in Figure~\ref{fig:mindmap}.
While there are concurrently related surveys in this domain, such as \citep{zhu2023survey} on LLM compression and \citep{miao2023towards}, \citep{ding2023efficiency} and \citep{wang2024model} on holistic LLM serving, our work stands out by incorporating a Roofline model analysis. 

In this paper, we first discuss the foundations of LLMs and develop a tool named LLM-Viewer, which uses Roofline model to analyze the bottleneck of deploying LLMs (Sec.\ref{sec:challenges}). 
LLM-Viewer can be used to analyze the deployment of any LLM architecture on various hardware platform, as shown in Figure \ref{fig:llmviewer_workflow}. 
For the literature review, this survey categorizes strategies for improving LLM inference efficiency into four main areas: Model Compression (Sec.\ref{sec:compression}), Algorithmic Methods for Fast Decoding (Sec.\ref{sec:fast-decoding}), Compiler/System-Level Optimization (Sec.\ref{sec:system_optimization}), and Hardware-Level Optimization (Sec.\ref{sec:hardware-optimization}).

%% file: sections/2_challenges.tex
\section{Delve into LLM Inference and Deployment}
\label{sec:challenges}



\subsection{LLM Inference}
\label{sec:llm_inference}


\begin{figure*}[t]
    \centering
    \includegraphics[width=0.8\linewidth]{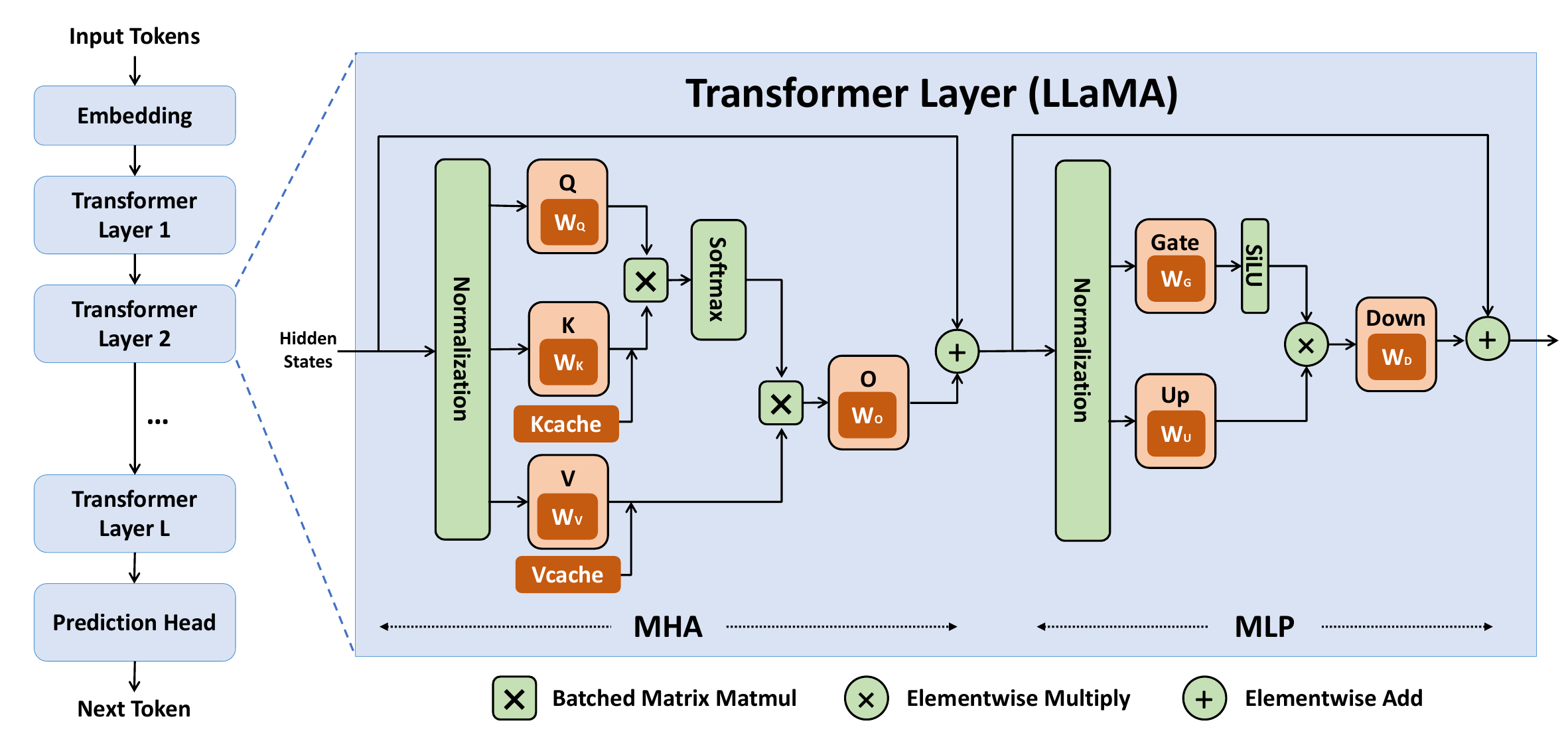}
    \caption{Demonstration of the architecture of LLMs.}
    \label{fig:transformer_layer}
\end{figure*}

Nowadays, the prevailing architecture adopted by most large language models (LLMs) is the Transformer decoder architecture. 
Here we will provide a concise overview of its fundamental structure, with the option to refer to this survey~\cite{zhao2023survey} for a more in-depth understanding.
This structure comprises an embedding layer, a series of sequential Transformer layers, and a prediction head. Figure~\ref{fig:transformer_layer} demonstrated the architecture. 


The embedding layer transform input tokens into the hidden states. 
The hidden states are sent to the Transformer layers.
Each Transformer layer consists of two components. 
Firstly, there is a masked multi-head attention module, denoted as \textbf{MHA}. 
Following MHA is a multi-layer perceptron submodule, labeled as \textbf{MLP}. 
The output from the last Transformer layer is then sent to the prediction head, which is responsible for predicting the next token after the input tokens.

Inference represents the process opposite to the training process. During training, a model learns from a vast dataset to capture the intricacies of language and context. The weights in model are updated. In contrast, during inference, a user inputs a \textbf{prompt}, and the LLM engages in a process of generating \textbf{responses}.
This process involves the model utilizing its fixed pre-trained weights to comprehend the input text and produce text as output. 
The inference process of Large Language Models (LLMs) is divided into two stages: the \textbf{Prefill Stage} and the \textbf{Decode Stage}.


The Prefill Stage serves as the initial step in LLM inference. In this stage, the model takes a prompt sequence as input and engages in the generation of a key-value cache (KV cache) for each Transformer layer within the LLM. The KV cache plays a crucial role in storing and organizing information that the model deems relevant for subsequent token generation. Each Transformer layer is equipped with its own unique KV cache, and this prefilling process establishes the foundation for the subsequent decoding stage.

In the Prefill Stage, the Multi-Head Attention (MHA) creats key-value (KV) pairs that will be stored in the KV cache. Let's denote the input to a Transformer layer as $\mathbf{X}_{\text{pre}} \in \mathbb{R}^{n\times d}$, where $d$ is the hidden size and $n$ is the length of prompt token sequence. The layers in the MHA have weights represented by $\mathbf{W}_q$, $\mathbf{W}_k$, $\mathbf{W}_v$, and $\mathbf{W}_o$.
The query, key and value are computed through the following process:
\begin{align*}
\text{Query:} \quad \mathbf{Q}_{\text{pre}} &=  \mathbf{X}_{\text{pre}} \cdot \mathbf{W}_q \\
\text{Key:} \quad \mathbf{K}_{\text{pre}} &=  \mathbf{X}_{\text{pre}} \cdot \mathbf{W}_k \\
\text{Value:} \quad \mathbf{V}_{\text{pre}} &= \mathbf{X}_{\text{pre}} \cdot \mathbf{W}_v 
\end{align*}
The generated $\mathbf{K}_{\text{pre}}$ and $\mathbf{V}_{\text{pre}}$ are stored in the KV cache. The other computation in MHA can be formulated as~\footnote{We omit layer norm, position mask, and positional embedding for simplicity.}:
\begin{align*}
\mathbf{O}_{\text{pre}} &= \text{softmax}\left(\frac{\mathbf{Q}_{\text{pre}} \cdot \mathbf{K}_{\text{pre}}^T}{\sqrt{d}}\right) \cdot \mathbf{V}_{\text{pre}}\cdot \mathbf{W}_o +\mathbf{X}_{\text{pre}},
\end{align*}
where the output of MHA $\mathbf{O}_{\text{pre}} \in \mathbb{R}^{n\times d}$ is sent to the MLP.
The output of the MLP serves as the input for the next Transformer layer.

The Decode Stage represents the core of the LLM inference process. In the Decode Stage, the model uses the KV caches prepared earlier and might add new information to them. The goal here is to generate tokens, which are essentially words or parts of words. This happens step by step. The creation of each new token is influenced by the tokens that were generated before it, like building a sentence word by word.

In the Decode Stage, the MHA loads the previously stored KV cache $\mathbf{K}_{\text{cache}}$ and $\mathbf{V}_{\text{cache}}$. The input is $\mathbf{X}_{\text{dec}} \in \mathbb{R}^{1\times d}$. New key and value pairs are computed and concatenated to the existing cache:
\begin{align*}
\text{Query:} \quad \mathbf{Q}_{\text{dec}} &=  \mathbf{X}_{\text{dec}} \cdot \mathbf{W}_q \\
\text{Key:} \quad \mathbf{K}_{\text{cat}} &=  [\mathbf{K}_{\text{cache}}, \mathbf{X}_{\text{dec}} \cdot \mathbf{W}_k] \\
\text{Value:} \quad \mathbf{V}_{\text{cat}} &=  [\mathbf{V}_{\text{cache}}, \mathbf{X}_{\text{dec}} \cdot \mathbf{W}_v]
\end{align*}
These newly computed $\mathbf{X}_{\text{dec}} \cdot \mathbf{W}_k$ and $\mathbf{X}_{\text{dec}} \cdot \mathbf{W}_v$ are then appended to the KV cache. The other computation in MHA is carried out as follows:
\begin{align*}
\mathbf{O}_{\text{dec}} &= \text{softmax}\left(\frac{\mathbf{Q}_{\text{dec}} \cdot \mathbf{K}_{\text{cat}}^T}{\sqrt{d}}\right) \cdot \mathbf{V}_{\text{cat}} \cdot \mathbf{W}_o + \mathbf{X}_{\text{dec}}
\end{align*}
where the output of MHA $\mathbf{O}_{\text{dec}} \in \mathbb{R}^{1\times d}$ is sent to the MLP.
The output of the last Transformer layer is sent to the final prediction layer to predict the next token.

\subsection{Roofline Model}
\label{sec:Rooflinemodel}




Assessing the efficiency at which LLMs deploy onto specific hardware involves a comprehensive consideration of both hardware and model characteristics. To conduct this evaluation, we employ the Roofline model.
The Roofline model serves as an effective theoretical framework to assess the potential performance of deploying a model on particular hardware. 

\begin{figure}[t]
    \centering
    \includegraphics[width=0.95\linewidth]{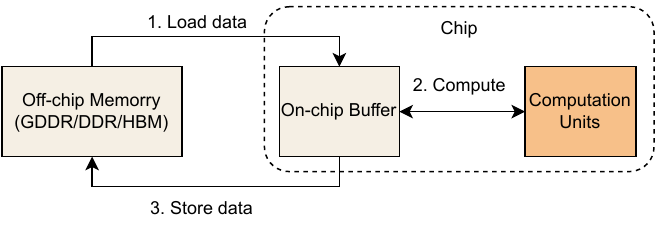}
    \caption{Execution of an operation on hardware.}
    \label{fig:hardware_run}
\end{figure}

As shown in Figure~\ref{fig:hardware_run}, the execution of a neural network layer on hardware devices entails the transfer of data from memory (DDR or HBM) to on-chip buffers, followed by computations performed by on-chip processing units, ultimately outputting results back to memory. 
Therefore, evaluating performance requires simultaneous consideration of memory access and processing unit capabilities.
If a layer involves extensive computations but minimal memory access, it is termed a computation bottleneck. 
This scenario leads to idle on the memory access. 
On the contrary, when a layer requires substantial memory access with fewer computational demands, it is referred to as a memory bottleneck. In this case, computational units remain underutilized. 
We can clearly distinguish between these two scenarios according to the Roofline model and provide performance upper bounds for different situations.

\begin{figure}[t]
    \centering
    \includegraphics[width=0.95\linewidth]{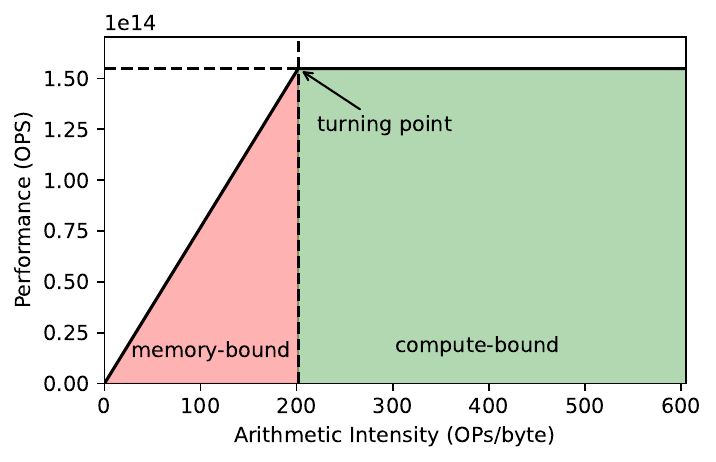}
    \caption{Demonstration of the Roofline model of Nvidia A6000 GPU. The computation is in FP16.}
    \label{fig:Roofline}
\end{figure}

There are two steps to using the Roofline model:

1. \textbf{Plot the Roofline}: Determine the peak computational performance (operations per second, OPS) and peak memory bandwidth (bytes per second) specific to the target hardware device.\footnote{OPS refer to the number of operations per second. Each Multiply-Accumulate (MAC) operation counts two operations.}
Then create a graph with performance (OPS) on the y-axis and arithmetic intensity (OPs/byte) on the x-axis: Draw a horizontal line equal to the peak computational performance. This line represents the maximum achievable performance by the hardware device.
And draw a diagonal line from the origin with a slope equal to the peak memory bandwidth. This line represents the maximum memory bandwidth available
on the system, known as the memory Roofline.
Figure~\ref{fig:Roofline} demonstrates the Roofline model of Nvidia A6000 GPU.

2. \textbf{Analyze performance for layers}: Evaluate the performance of each layer in the model by quantifying both the number of operations (OPs) and the volume of data accessed from memory (bytes). Calculate the arithmetic intensity (OPs/byte) of each layer by dividing the required operations by the amount of data transferred. 
According to the graph created in the first step, the theoretical max performance for each layer is determined by the position on the graph corresponding to the x-axis value of arithmetic intensity. It allows us to ascertain whether the system is memory-bound or compute-bound at this point, guiding the determination of the subsequent optimization strategy.

\begin{table}[tb]
\caption{Analysis for layers in Llama-2-7b using the Roofline model of Nvidia A6000 GPU. In this example, the sequence length is 2048 and the batch size is 1.} 
\label{tab:Llama_analysis} 
\resizebox{\linewidth}{!}{
\begin{tabular}{@{}cccccc@{}}
\toprule
Layer Name & OPs  & \begin{tabular}[c]{@{}c@{}}Memory\\ Access\end{tabular} & \begin{tabular}[c]{@{}c@{}}Arithmetic\\ Intensity\end{tabular} & \begin{tabular}[c]{@{}c@{}}Max\\ Performance\end{tabular} & Bound   \\ \midrule
\multicolumn{6}{c}{Prefill}                                                                                                                                                                                        \\ \midrule
q\_proj    & 69G  & 67M                                                     & 1024                                                           & 155T                                                      & compute \\
k\_proj    & 69G  & 67M                                                     & 1024                                                           & 155T                                                      & compute \\
v\_proj    & 69G  & 67M                                                     & 1024                                                           & 155T                                                      & compute \\
o\_proj    & 69G  & 67M                                                     & 1024                                                           & 155T                                                      & compute \\
gate\_proj & 185G & 152M                                                    & 1215                                                           & 155T                                                      & compute \\
up\_proj   & 185G & 152M                                                    & 1215                                                           & 155T                                                      & compute \\
down\_proj & 185G & 152M                                                    & 1215                                                           & 155T                                                      & compute \\
qk\_matmul & 34G  & 302M                                                    & 114                                                            & 87T                                                       & memory  \\
sv\_matmul & 34G  & 302M                                                    & 114                                                            & 87T                                                       & memory  \\
softmax    & 671M & 537M                                                    & 1.25                                                           & 960G                                                      & memory  \\
norm       & 59M  & 34M                                                     & 1.75                                                           & 1T                                                        & memory  \\
add        & 8M   & 34M                                                     & 0.25                                                           & 192G                                                      & memory  \\ \midrule
\multicolumn{6}{c}{Decode}                                                                                                                                                                                         \\ \midrule
q\_proj    & 34M  & 34M                                                     & 1                                                              & 768G                                                      & memory  \\
k\_proj    & 34M  & 34M                                                     & 1                                                              & 768G                                                      & memory  \\
v\_proj    & 34M  & 34M                                                     & 1                                                              & 768G                                                      & memory  \\
o\_proj    & 34M  & 34M                                                     & 1                                                              & 768G                                                      & memory  \\
gate\_proj & 90M  & 90M                                                     & 1                                                              & 768G                                                      & memory  \\
up\_proj   & 90M  & 90M                                                     & 1                                                              & 768G                                                      & memory  \\
down\_proj & 90M  & 90M                                                     & 1                                                              & 768G                                                      & memory  \\
qk\_matmul & 17M  & 17M                                                     & 0.99                                                           & 762G                                                      & memory  \\
sv\_matmul & 17M  & 17M                                                     & 0.99                                                           & 762G                                                      & memory  \\
softmax    & 328K & 262K                                                    & 1.25                                                           & 960G                                                      & memory  \\
norm       & 29K  & 16K                                                     & 1.75                                                           & 1T                                                        & memory  \\
add        & 4K   & 16K                                                     & 0.25                                                           & 192G                                                      & memory  \\ \bottomrule
\end{tabular}
}
\end{table}

There are two scenarios where resources are not fully utilized: When the model's computational intensity is below the turning point, residing in the red zone, it implies that the computational workload required per memory access is low. Even saturating the peak bandwidth does not fully utilize all computational resources. In such cases, the layer is constrained by memory access (memory-bound), and some computational units may remain idle.
If the layer is memory-bound, consider optimization techniques such as quantization, kernel fusion and increasing batch size to alleviate the memory footprint. 
Conversely, if the model's computational intensity is above the turning point, situated in the green zone, it suggests that the model requires only a small amount of memory access to consume a significant amount of computational capability. It implies that the layer is constrained by computation (compute-bound), with some memory units potentially remaining idle.
In this case, we should investigate strategies such as enabling low-bit computation to enhance computational efficiency. 
Detailed explanations of these methods will be provided in the subsequent sections.


As an example, Table~\ref{tab:Llama_analysis} presents the analysis of layers in Llama-2-7b using the Roofline model on the Nvidia A6000 GPU.
From the table, we observe that during the prefill stage, the majority of computations are compute-bound, leading to high performance. 
Conversely, in the decode stage, all computations are memory-bound, resulting in performance significantly below the computational capacity of the GPU's computation units.
During the user's interaction with large models, the prefill stage executes only once, while the decode stage is repeatedly performed to generate a continuous output. Therefore, optimizing for the memory-bound characteristics of the decode stage becomes crucial for enhancing the inference performance of large models.

\subsection{LLM-Viewer}
\label{sec:llmviewer}


There are multiple Transformer layers in LLMs, each containing various operations. Moreover, different LLMs have different sets of operations. Additionally, we need to track information like memory footprint to calculate the peak memory usage and total inference time. Hence, analyzing LLMs involves examining network-wide concerns.
In this section, we propose a powerful tool, LLM-Viewer~\footnote{The tool is open-sourced at https://github.com/hahnyuan/LLM-Viewer}, to execute the network-wise analysis.
It enables the analysis of LLM performance and efficiency on various hardware platforms, offering valuable insights into LLM inference and performance optimization.

The workflow of LLM-Viewer is depicted in Figure~\ref{fig:llmviewer_workflow}. It consists of the following steps:
(1) Input the LLM and gather essential information about each layer, such as the computation count, input and output tensor shapes, and data dependencies.
(2) Provide input for the hardware and generate a Roofline model that takes into account the computation capacity and memory bandwidth of the hardware.
(3) Configure the inference settings, including the batch size, prompt token length, and generation token length.
(4) Configure the optimization settings, such as the quantization bitwidth, utilization of FlashAttention, decoding methods, and other system optimization techniques.
(5) The LLM-Viewer Analyzer utilizes the Roofline model and layer information to analyze the performance of each layer. It also tracks the memory usage of each layer and calculates the peak memory consumption based on data dependencies. By aggregating the results of all layers, the overall network performance of LLM can be obtained.
(6) Generate a report that provides information such as the maximum performance and performance bottlenecks of each layer and the network, as well as the memory footprint. Analyzing curves, such as batch size-performance and sequence length-performance curves, can be plotted from the report to understand how different settings impact performance.
(7) LLM-Viewer offers a web viewer that allows convenient visualization of the network architecture and analysis results. This tool facilitates easy configuration adjustment and provides access to various data for each layer.

%% file: sections/3_compression.tex
\section{Model Compression}
\label{sec:compression}
The formidable size and computational demands of Large Language Models (LLMs) present significant challenges for practical deployment, especially in resource-constrained environments. To alleviate these limitations, the most straightforward solution is to compress the LLMs. 
In this section, we review the concept of neural network compression for LLMs. This exploration encompasses a thorough examination of well-established techniques, including but not limited to quantization, pruning, knowledge distillation, and low-rank factorization.
In each subsection, we will utilize LLM-Viewer to analyze the impact of network compression on LLM inference. Based on our analysis, we will provide optimization recommendations.

\subsection{Quantization}
\label{sec:quantization}
In the realm of LLM compression, quantization has become a pivotal technique for mitigating the substantial storage and computational overhead associated with these models. Essentially, quantization involves transforming the floating-point values in original LLMs into integers or other discrete forms, a process that considerably reduces both storage requirements and computational complexity~\citep{gholami2022survey}. While some degree of precision loss is inherent in this process, carefully designed quantization techniques can achieve significant model compression with minimal impact on accuracy.
Quantization in the context of LLMs can be primarily categorized into two directions: Quantization for Compressing Pre-trained LLMs and Quantization for Parameter-Efficient Fine-Tuning (Q-PEFT). 
The first category encompasses approaches that apply quantization to LLMs for using the quantized LLMs as pre-trained models. This category can be further divided into two subcategories: Quantization-Aware Training (QAT) and Post-Training Quantization (PTQ). 
QAT integrates quantization into the model’s training process or during the fine-tuning/re-training of a pre-trained LLM, allowing the model to adapt to the quantization from the onset. In contrast, PTQ applies quantization to a model after it has completed its training phase, offering a more straightforward approach to model compression without the need for retraining. These distinct methodologies highlight the versatility of quantization techniques in addressing the specific needs and constraints of LLM deployment.

\subsubsection{A Use Case of LLM-Viewer: \\Roofline Analysis for Quantization}
Here we provide an example of how to use our LLM-Viewer (Section~\ref{sec:llmviewer}) to analyze the bottlenecks of LLM deployments. 
In LLMs, tensors consist of weights and activations, with activations including temporary activations and KV cache.
(1) LLM weights must be stored in memory. For example, Llama-13b~\citep{touvron2023llama1}, which has 13 billion weights, occupies approximately 26GB of memory in FP16 format. 
(2) temporary activations are generated during inference. For example, the inputs of each transformer layer are kept in memory until the residual addition is executed.
(3) for auto-regressive LLMs, caching key and value activations (KV cache) into memory is necessary for subsequent token generation.
We utilize LLM-Viewer to analyze the effects of quantization on these tensors from three perspectives: computation, memory consumption, and memory access.

\begin{figure}[t]
    \centering
    \includegraphics[width=0.95\linewidth]{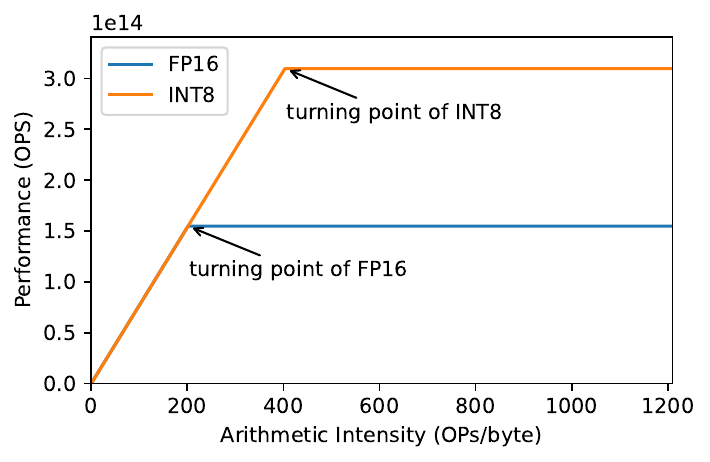}
    \caption{Demonstration of the Roofline model of Nvidia A6000 GPU for different computation data types.}
    \label{fig:quantization_roofline_model}
\end{figure}

\begin{figure}[t]
    \centering
    \includegraphics[width=0.95\linewidth]{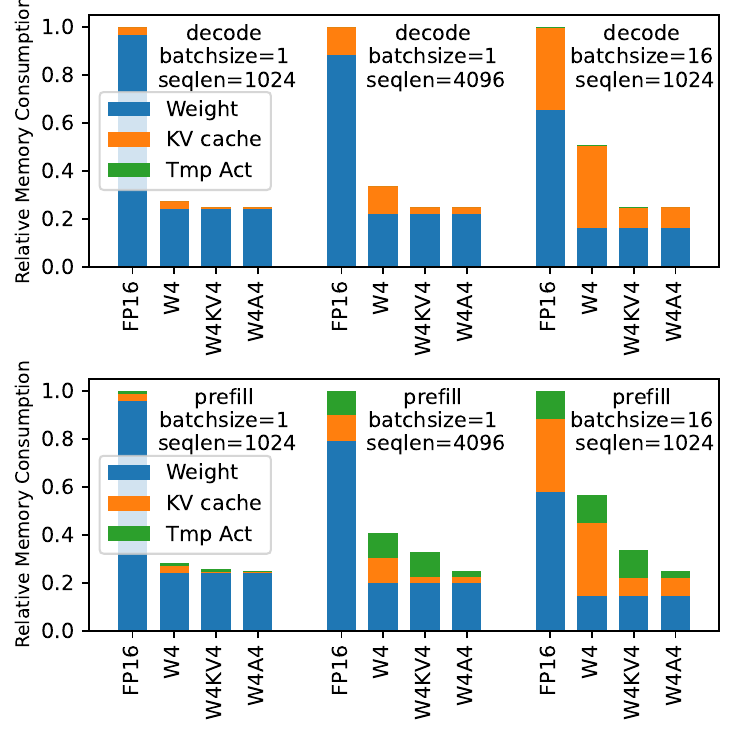}
    \caption{Relative memory consumption for different quantization settings for Llama-2-13b. Tmp Act means temporary activations.}
    \label{fig:quantization_memory_consumption}
\end{figure}

\textbf{Computation}: 
The latest computing devices, such as NVIDIA GPUs, generally support FP32, FP16, and INT8 data types for computation. Hardware devices generally perform better when processing data with smaller bit widths. NVIDIA's A6000 GPU, for example, is capable of performing twice as fast as FP16 with 155 TOP/s and 310 TOP/s, respectively.
In the Roofline model, when enabling quantization for faster computation, the roofline height increases, indicating improved performance for compute-bound layers. 
As shown in Figure~\ref{fig:quantization_roofline_model}, the max performance improved when using INT8 computation.
However, to utilize the computational power of INT8, all input operands must be in INT8 format. Consequently, if only the weights are quantized to INT8 while the activations remain in FP16 format, the INT8 computational power cannot be utilized. Instead, the INT8 weights would need to be converted to FP16 for multiplication with FP16 activations.
Furthermore, when tensors are quantized to a bitwidth that is not supported by the hardware, they need to be converted to higher bit widths for computation. For example, the NVIDIA H100 GPU does not support INT4 computation. Consequently, if the weight or activation is quantized to INT4, it would require conversion to a higher bit width, such as INT8 or FP16, for computation.

\textbf{Memory Consumption}: The memory consumption reduction resulting from quantizing different tensors varies, as shown in Figure~\ref{fig:quantization_memory_consumption}~\footnote{In our notation, W4 represents the quantization of weights to 4 bits while keeping activations in FP16 format. W4A4 indicates both weights and activations quantized to 4 bits. In the case of W4KV4, weights and the KV cache are quantized, while temporary activations remain in FP16 format.}. Notably, the memory usage of temporary activations is relatively low, especially during the decode stage. This can be attributed to their short lifespan, allowing their memory to be released once their purpose is fulfilled.
On the other hand, the memory allocated for the KV cache behaves differently. It cannot be freed until the entire process of generating a complete answer is finished, which entails multiple inference passes through the network. Additionally, the memory consumption of the KV cache increases as the batch sizes grow larger and the input sequences become longer. This is because the model needs to store a greater number of key-value (KV) pairs to facilitate its operations.

\textbf{Memory Access}: Quantizing tensors in LLM can significantly reduce memory access, resulting in fewer data bytes to be moved for the same amount of computation. This increase in arithmetic intensity contributes to the Roofline model, leading to three scenarios:
(1) After quantization, the arithmetic intensity remains within the memory-bound range. With the improvement in arithmetic intensity, the average data access per computation is reduced, alleviating the pressure on data memory access. Consequently, the theoretical performance is enhanced. This can greatly boost the performance during the memory-bound decode stage.
(2) The arithmetic intensity transitions from being memory-bound to compute-bound. This shift also reduces the pressure on data memory access, resulting in improved theoretical performance.
(3) Both before and after quantization, the arithmetic intensity remains within the compute-bound range. In this case, there is no performance improvement. For example, this scenario can occur during the compute-bound prefill stage or when the batch size is large in the decode stage.

\begin{figure}[t]
    \centering
    \includegraphics[width=0.95\linewidth]{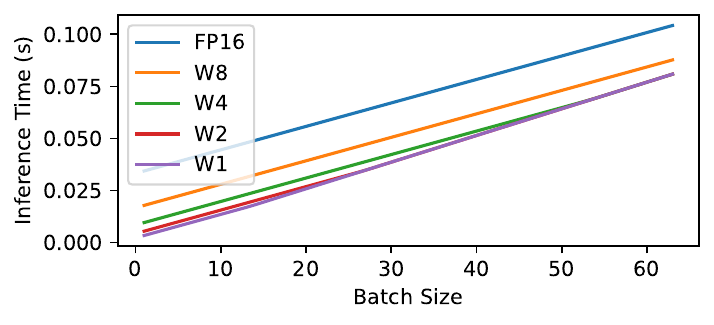}
    \caption{Inference time of decoding stage for different quantization settings on Llama-2-13b. (Sequence length=1024)}
    \label{fig:quantization_memory_access_batch}
\end{figure}

\begin{figure}[t]
    \centering
    \includegraphics[width=0.95\linewidth]{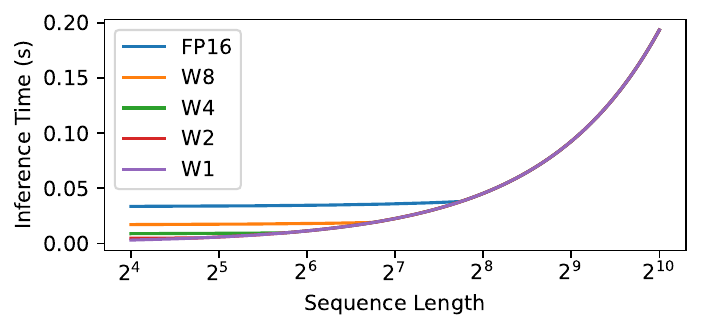}
    \caption{Inference time of prefill stage for different quantization settings on Llama-2-13b. (Batch size=1)}
    \label{fig:quantization_memory_access_seq_len}
\end{figure}

As depicted in Figure~\ref{fig:quantization_memory_access_batch}, when the batch size is small, the layers in the network are memory-bound both before and after quantization. Therefore, quantization can enhance performance and reduce the network's inference time. However, when the batch size is large, compressing the network's weights from 4 bits to 2 bits or 1 bit does not lead to a decrease in the inference time. This is because, at this point, the network is already compute-bound, and quantizing the weights becomes ineffective.
Similar to the previous scenario, the behavior of the system can exhibit saturation effects in prefill stage. As shown in Figure~\ref{fig:quantization_memory_access_seq_len}, when the sequence length is relatively small, the prefill stage is memory-bound. In this case, applying quantization can enhance the performance by reducing the memory access requirements of the network.
However, as the sequence length increases, the prefill stage becomes more compute-bound. Consequently, quantizing the weights may not yield significant improvements in performance when the network is already compute-bound during the prefill stage with large sequence lengths.


\subsubsection{Quantization for Compressing Pre-trained LLMs}
In Quantization-Aware Training (QAT)~\citep{courbariaux2015binaryconnect,choi2018pact,dong2019hawq}, the quantization process is seamlessly integrated into the training of Large Language Models (LLMs), enabling them to adapt to low-precision representations and thus mitigating precision loss. LLM-QAT~\citep{liu2023llm} innovatively addresses the challenge of training data acquisition for LLMs through data-free distillation, which leverages outputs from a pre-trained model to obviate the need for extensive data collection. Furthermore, LLM-QAT expands quantization beyond weights and activations to include key value (KV) caches, enhancing throughput and supporting longer sequence dependencies. Its successful distillation of large Llama models to 4-bit quantized weights and KV caches underscores the potential for accurately quantized 4-bit LLMs.

To attain lower-bit quantization, such as below 2-bit, \cite{kim2023token} introduce Token-Scaled Logit Distillation (TSLD) for ternary QAT in LLMs. This method employs an adaptive knowledge distillation technique that modifies Logit Knowledge Distillation based on token confidence, providing tailored guidance during LLM QAT. Furthermore, \cite{shang2024pb} focus on salient weights with their concept of partially binarized matrices in PB-LLM. By preserving these crucial weights in higher bits, PB-LLM effectively maintains the reasoning capacity of heavily quantized LLMs. Additionally, PB-LLM explores minimizing quantization error by determining the optimal scaling factors for binarized LLMs, a vital step in preserving the effectiveness of models under aggressive quantization.

\begin{figure*}[t]
    \centering
    \includegraphics[width=0.99\linewidth]{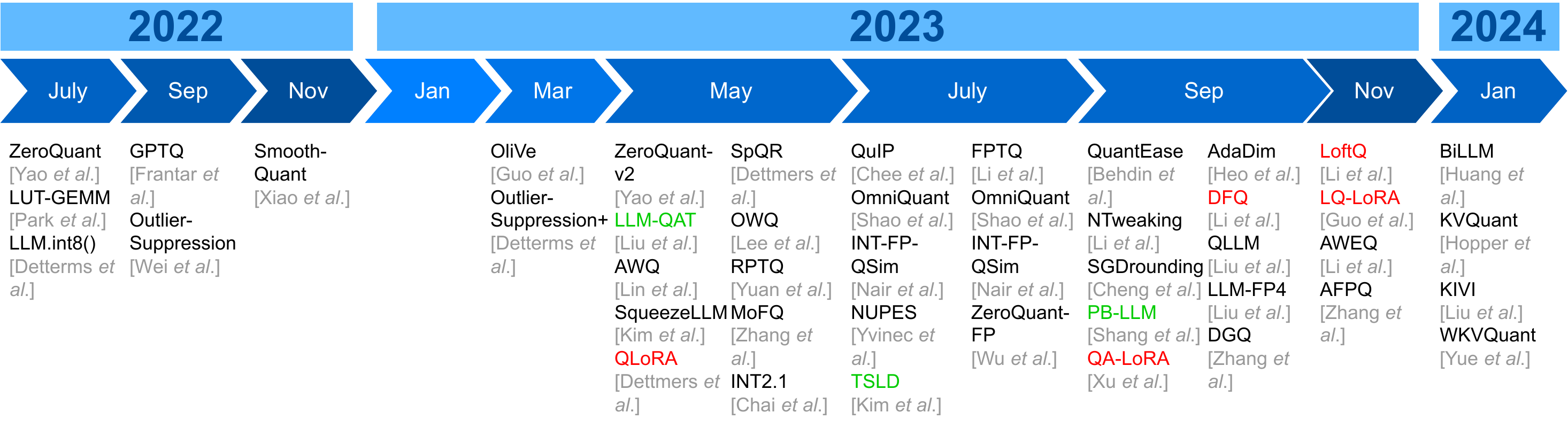}
    \vspace{-0.1in}
    \caption{Timeline of Quantization for LLM methods from 2022 to 2024. The \textcolor{red}{red}-highlighted methods represent they belonging to Quantization for Parameter Efficient Fine-Tuning (Q-PEFT), the \textcolor{green}{green}-highlighted methods represent they belonging to QAT-related methods, and others are PTQ-based methods.}
    \label{fig:ptq-timeline}
\end{figure*}
 
\noindent\textbf{Post-Training Quantization (PTQ)}

Post-Training Quantization (PTQ) represents a crucial technique in optimizing Large Language Models (LLMs), entailing the quantization of model parameters post the LLM's training phase. The primary goal of PTQ is to reduce both the storage requirements and computational complexity of the LLM, without necessitating alterations to the model's architecture or embarking on a retraining process. This approach stands out for its simplicity and efficiency, particularly in achieving significant model compression. In the context of LLMs, which typically contain billions of parameters, Quantization-Aware Training (QAT) often becomes impractical due to excessive training costs. Hence, PTQ emerges as a more viable solution for these large-scale models. However, it's crucial to acknowledge that PTQ can lead to a certain degree of precision loss as a consequence of the quantization process. Despite this, PTQ serves as an effective method to enhance the efficiency of LLMs, offering a straightforward solution that avoids major modifications or extensive additional training.

In PTQ, various approaches focus on \textbf{weight-only quantization} to enhance efficiency. For instance, LUT-GEMM~\citep{park2023lut} optimizes matrix multiplications in LLMs using weight-only quantization and the BCQ format, thereby reducing latency and improving computational efficiency. 
 LLM.int8()~\citep{dettmers2022llm} employs 8-bit quantization, which halves GPU memory usage during inference and maintains precision through vector-wise quantization and mixed-precision decomposition. This method enables efficient inference in models up to 175 billion parameters. 
ZeroQuant~\citep{yao2022zeroquant} combines a hardware-friendly quantization scheme with layer-by-layer knowledge distillation, optimizing both weights and activations to INT8 with minimal accuracy loss. 
Addressing higher compression targets, GPTQ~\citep{frantar2022gptq} introduces a layer-wise quantization technique based on approximate second-order information, achieving a reduction to 3-4 bits per weight with minimal accuracy loss. Additionally, the study by \cite{dettmers2023case} explores the balance between model size and bit precision, particularly for zero-shot performance, finding that 4-bit precision generally offers the optimal balance.
Innovations like AWQ~\citep{lin2023awq, kim2023squeezellm} highlight that protecting a small percentage of salient weights can significantly reduce quantization error. AWQ uses an activation-aware approach, focusing on weight channels with larger activation magnitudes, and incorporates per-channel scaling for optimal quantization. OWQ~\citep{lee2023owq} analyzes how activation outliers amplify quantization error, introducing a mixed-precision scheme to assign higher precision to weights affected by these outliers. SpQR~\citep{dettmers2023spqr} takes a unique approach by isolating outlier weights for storage in higher precision, while compressing the remainder to 3-4 bits. This technique allows for more efficient compression while maintaining near-lossless performance.
QuantEase~\citep{behdin2023quantease} suggests using a coordinate descent approach to optimize all of the weights in network, improving the efficiency of quantization.

To achieve even lower-bit quantization (e.g., below 2-bit), QuIP~\citep{chee2023quip} introduces an innovative approach that accounts for the even distribution of weight magnitudes and the significance of accurately rounding directions unaligned with coordinate axes. QuIP comprises an adaptive rounding procedure that minimizes a quadratic proxy objective, essential for optimizing the quantization process. Additionally, it employs efficient pre- and post-processing techniques that ensure weight and Hessian incoherence through multiplication by random orthogonal matrices, crucial for maintaining quantization effectiveness.
Further advancing PTQ methods, \cite{li2023norm} are inspired by the observation that aligning the quantized activation distribution with its floating-point counterpart can restore accuracy in LLMs. Their proposed 'Norm Tweaking' strategy involves a meticulous calibration data generation process and a channel-wise distance constraint. This approach updates the weights of normalization layers, leading to enhanced generalization capabilities.
\citep{shang2024pb}~propose partial-binarized LLM (PB-LLM) by introducing binarization~\citep{hubara2016binarized}~into LLM quantization to push weight quantization under 2 bits. Following PB-LLM, BiLLM~\citep{huang2024billm} pushes weight quantization to almost 1 bit.


Apart from efforts that focus solely on weight quantization in LLMs, numerous PTQ approaches focus on \textbf{weights and activations quantization}. SmoothQuant~\citep{xiao2023smoothquant} addresses the challenge of quantizing activations, which can be complex due to the presence of outliers. It introduces a per-channel scaling transformation that effectively smooths out activation magnitudes, rendering the model more receptive to quantization.
Recognizing the intricacies of quantizing activations in LLMs, RPTQ~\citep{yuan2023rptq} highlights the uneven ranges across channels and the prevalence of outliers. RPTQ's innovative approach involves clustering channels for quantization, thereby reducing discrepancies in channel ranges. This method smartly integrates channel reordering into layer normalization and linear layer weights to minimize overhead.
OliVe~\citep{guo2023olive} adopts an outlier-victim pair (OVP) quantization strategy, focusing on local handling of outliers with low hardware overhead and significant performance benefits. This approach stems from the understanding that outliers are crucial, while adjacent normal values are less so. Building on this, Outlier Suppression+ extends the concept by addressing asymmetrically distributed harmful outliers in specific channels. It introduces channel-wise shifting and scaling operations to balance the outlier distribution and reduce the impact of problematic channels, considering both the nature of the outliers and the subsequent quantization errors.
ZeroQuant-FP~\citep{wu2023zeroquant} delves into floating-point (FP) quantization, specifically exploring FP8 and FP4 formats. This study finds that FP8 activation quantization in LLMs outperforms the traditional INT8 format, while FP4 weight quantization shows comparable efficacy to INT4. ZeroQuant-FP addresses the divergence between weights and activations by standardizing all scaling factors as powers of 2 and restricting them within a single compute group, ensuring consistency and efficiency in the quantization process.
\cite{li2023fptq} propose FPTQ, in which they employ a layerwise strategy to cope with different levels of quantization difficulty. Particularly, they devises an offline logarithmic activation equalization to render a quantization-friendly distribution for previously intractable layers.

\begin{figure}[t]
    \centering
    \includegraphics[width=0.95\linewidth]{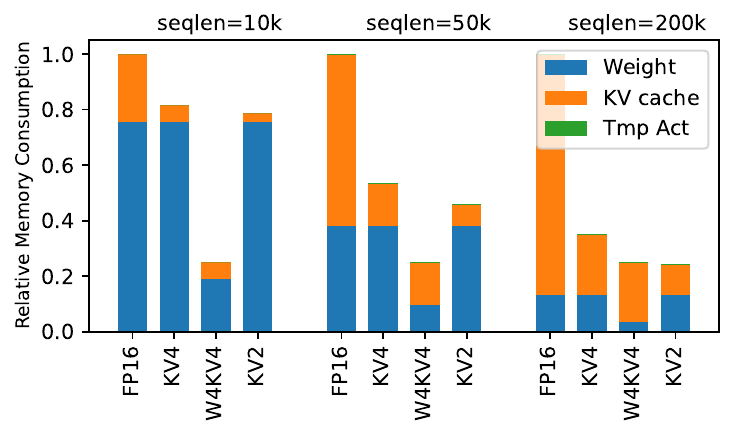}
    \caption{Memory Consumption of decode stage for different quantization settings on Llama-2-13b. (Batch size=1).}
    \label{fig:quantization_memory_consumption_long_seq}
\end{figure}

Since the end of 2023, the length of tokens has been significantly increasing, causing the KV cache to consume more memory. 
For instance, Google Gemini 1.5~\citep{gemini2024} can handle up to 1 million tokens in production, and LLMs processing books, large images or videos will require tens of thousands of tokens. As a result, the optimization of \textbf{KV Cache Quantization} has become increasingly important. 
Several recent papers in 2024 have focused on improving KV cache quantization. 
For example, \cite{hooper2024kvquant} propose a solution for achieving 10 million context length LLM inference with KV Cache Quantization.
KIVI~\citep{liu2024kivi} pushes the quantization of KV cache to 2-bit.
\cite{yue2024wkvquant} proposes WKVQuant as to jointly optimize the quantization of both the weights and the KV cache in LLMs, making W4KV4 have the same performance as W4.
As shown in Figure~\ref{fig:quantization_memory_consumption_long_seq}, we use LLM-Viewer to analyze the memory reduction of KV cache quantization.
We can observe that when the sequence length is larger than 50k, the KV cache takes most of the memory and its quantization can significantly decrease the memory consumption.


\subsubsection{Quantization for Parameter Efficient Fine-Tuning (Q-PEFT)}

\textbf{Parameter Efficient Fine-Tuning (PEFT)} is an important topic for LLMs. One of the most popular approaches is low-rank adaptation (LoRA)~\citep{hu2021lora,valipour2022dylora}, where the key insight is to decompose the adapter weights into the multiplication of two low-rank (and thus parameter-efficient) matrices. 
LoRA has claimed comparable performance to full fine-tuning while using much fewer learnable parameters. 
Please refer to the review paper~\citep{hu2023llm} for more details about this adaptor. 

In addition to the well-defined quantization paradigms, a novel paradigm in LLM efficiency is emerging: Quantization for Parameter-Efficient Fine-Tuning (Q-PEFT).
This approach integrates quantization into the fine-tuning process of LLMs, offering a unique and efficient method, particularly relevant in the era of large models. 
Pioneering works in this paradigm, such as PEQA~\citep{kim2023memory}, DFT~\citep{li2023qft}, and QLORA~\citep{dettmers2023qlora} demonstrate the feasibility and effectiveness of this approach. 
PEQA employs a dual-stage process where the first stage involves quantizing the parameter matrix of each fully connected layer into a matrix of low-bit integers coupled with a scalar vector. The second stage focuses on fine-tuning the scalar vector for specific downstream tasks, allowing for more efficient task-specific adjustments. 
DFT adopts the efficient Lion optimizer, which only keeps track of the momentum and has consistent update magnitudes for each parameter, an inherent advantage for robust quantization; and (ii) we quantize all model states and store them as integer values, and present a gradient flow and parameter update scheme for the quantized weights. 
On the other hand, QLORA introduces novel concepts such as a new data type, double quantization, and paged optimizers. These innovations aim to conserve memory efficiently while maintaining LLM fine-tuning performance. Notably, QLORA facilitates large model fine-tuning on a single GPU, achieving state-of-the-art results on the Vicuna benchmark, a testament to its effectiveness in balancing memory efficiency and model performance.

However, a limitation of QLoRA is its restriction to at most 4-bit quantization during fine-tuning; lower-bit quantization, such as 2-bit, can significantly deteriorate the performance. Addressing this challenge, several studies have ventured into the realm of Q-PEFT to enable lower-bit quantization. LQ-LoRA~\citep{guo2023lq} introduces an iterative algorithm that decomposes each pretrained matrix into a high-precision, low-rank component and a memory-efficient quantized component. During fine-tuning, only the low-rank component is updated, keeping the quantized component fixed. This method presents an integer linear programming approach for the quantization component, allowing dynamic configuration of quantization parameters like bit-width and block size within a given memory budget. Another notable approach, Loft-Q~\citep{li2023loftq}, simultaneously quantizes an LLM and establishes a suitable low-rank initialization for LoRA fine-tuning. This strategy effectively bridges the gap between the quantized and full-precision models, significantly enhancing generalization in downstream tasks. 
QA-LoRA~\citep{xu2023qa} leverages the benefits of quantizing the LLM’s weights into low-bit integers, facilitating an efficient fine-tuning stage. Additionally, it produces a lightweight, fine-tuned model, circumventing the accuracy loss often associated with PTQ.

\subsubsection{Discussion on LLM Quantiztaion}
Figure~\ref{fig:ptq-timeline} presents a timeline of LLM quantization techniques, highlighting the evolution from Post-Training Quantization (PTQ) as the initial mainstream approach to the rising prominence of Quantization-Aware Training (QAT) and Quantization for Parameter-Efficient Fine-Tuning (Q-PEFT). This shift underscores the community's adaptation in response to the performance bottlenecks encountered with PTQ, marking QAT and Q-PEFT as the burgeoning areas of focus in the quest for efficient LLM inference.

\subsection{Pruning}
Pruning~\citep{lecun1989optimal,liang2021pruning}, which concentrates on identifying and eliminating model parameters that are deemed unnecessary or redundant, is another popular technique for compressing LLMs.
In the context of LLMs, the parameters often account for a considerable portion of the model size and computational demand. By carefully pruning these parameters, it's possible to streamline the model, making it more efficient without significantly compromising its performance. 
Pruning methods can be broadly classified into two categories: unstructured pruning and structured pruning, and we describe the research progress of each category in turn below.

\subsubsection{Unstructured pruning} 
Unstructured pruning selectively eliminates individual weights or neurons from a model, leading to a sparser, yet more irregularly structured network. This form of pruning excels in ensuring model accuracy, however, the resultant irregularity in the weight distribution necessitates specialized handling or software optimizations.
SparseGPT~\citep{frantar2023sparsegpt} is a groundbreaking one-shot pruning method tailored for LLMs. It tackles the pruning challenge by reconceptualizing it into a series of extensive sparse regression problems, efficiently solved by a newly developed solver. Notably, SparseGPT can efficiently process a model with 175 billion parameters in just a few hours on a single GPU, and it can induce significant sparsity (50-60\%) in LLMs without significantly sacrificing accuracy or necessitating fine-tuning. 
To tackle the challenge of reconstruction cost in SparseGPT, \cite{sun2023simple} propose Wanda, which assesses the significance of each weight by evaluating its magnitude and the norm of the corresponding input, significantly increasing the computational efficiency.
Further, \cite{yin2023outlier} design a set of non-uniform hierarchical sparsity ratios to pay more attention to the layers with higher outlier occurrences, thus boosting the pruning performance.
Moreover, Considering the hardware support for unstructured pruning, Flash-LLM~\citep{xia2023flash} proposes an unstructured sparse matrix multiplication method, which is characterized by sparse loading and dense computation, to implement the GPU Tensor Core's sophisticated support for unstructured sparsity.

\subsubsection{Structured pruning} 
Structured pruning removes entire neurons or layers, resulting in a cleaner, more regular structure. The pruned model is generally more compatible with conventional hardware, however, the simplicity and regularity come at a cost: this form of pruning can have a more pronounced impact on the model's performance, as it involves removing larger, potentially more critical components.
LLM-Pruner~\citep{ma2023llm} represents a pioneering approach in structural pruning for LLMs. It employs a one-shot pruning technique, which relies on first-order and estimated Hessian data and necessitates subsequent fine-tuning using LoRA to restore the weights. This work is advantageous as it significantly reduces both computational demands and memory requirements, while preserving the fundamental structure of LLMs. 
Sheared Llama~\citep{xia2023sheared} proposes another noteworthy solution by combining targeted structured pruning with a dynamic batch loading algorithm. First, it meticulously prunes a source model into a desired target architecture, meticulously chosen by analyzing the configurations of the pre-trained model. Then, it enhances training efficiency through the dynamic batch loading algorithm, which adjusts the proportion of training data from various domains.
Compresso~\citep{guo2023compresso} establishes a collaborative learning framework, where the LLM and a resource-efficient pruning algorithm work in tandem, with the ability to prune Llama-7B to 5.4B while preserving the original performance.

\subsection{Knowledge Distillation}
Knowledge distillation~\citep{hinton2015distilling,gou2021knowledge} is a technique that facilitates the transfer of capabilities from a larger model (referred to as the ``teacher'') to a smaller model (referred to as the ``student''), allowing the smaller model to perform tasks with similar proficiency as the larger model but with reduced computational resources~\citep{gou2021knowledge,shang2021lipschitz}. 
For LLM compression, there are two main categories of knowledge distillation: white-box and black-box distillation. Within these categories, researchers have developed a range of distillation methods tailored for LLMs, which are described in detail below. 
Moreover, a more detailed and specific survey regarding knowledge distillation of LLMs has also been carried out~\citep{xu2024kdsurvey}. 

\subsubsection{White-Box Knowledge Distillation}
In white-box distillation, the architecture and weights of the teacher model are fully accessible. This transparency allows the student model to learn not just the output of the teacher model but also its internal representations and decision-making processes.
MiniLLM~\citep{gu2023knowledge} critiques the limitations of standard knowledge distillation objectives and suggests that reverse Kullback-Leibler divergence is more effective for capturing the complexity of generative tasks, which can enhance the student model's response quality and reliability. MiniLLM also introduces single-step regularization, teacher-mixed sampling, and length normalization to address challenges in training, thus demonstrating great performance potential for distilling LLMs on the standard benchmarks.
In contrast to MiniLLM, GKD~\citep{agarwal2023gkd} presents a more straightforward and stable method. It aligns more with supervised training by avoiding backpropagation through the student model’s sampling. Instead of using predetermined output sequences, GKD trains the student model on its own created sequences, utilizing the teacher's probabilities as guidance, which leads to notable improvements in the student's performance.
Homotopic distillation~\citep{liang2023homodistil} aims to facilitate the alignment of the student model's predictions with those of the teacher model across extensive open-domain data. It involves starting the student model with the teacher model's configuration and progressively reducing the student model's neurons to reach a specified model complexity.
Furthermore, ~\cite{liang2023less} present a layerwise distillation approach that involves creating unique task-aware filters for each layer of teacher and student models. These filters, essentially neural networks equipped with task-specific heads, are designed to distill and capture the predictive knowledge from the hidden layers of the respective models.
AD-KD~\citep{wu2023ad} analyzes the teacher model's token-level rationale using integrated gradients and transfers attributional knowledge to the student model, which enables the student model to imitate the teacher's underlying reasoning, not just its behaviors.

\subsubsection{Black-Box Knowledge Distillation}
Contrary to white-box distillation, black-box distillation does not require access to the internal information of the teacher model. Instead, it focuses on replicating the output behavior of the teacher model. The student model learns solely from the input-output pairings produced by the teacher, without any insight into its internal operations.
Multitask-ICT~\citep{huang2022context} introduces in-context learning distillation, which merges the objectives of in-context learning with those of language modeling, intending to distill into smaller models both the capability to comprehend in-context examples and the knowledge required for specific tasks.
LaMini-LM~\citep{wu2023lamini} creates a set of 2.58 million instructions and employs GPT-3.5 Turbo to produce responses to these instructions. Subsequently, it uses these instructions as a basis to fine-tune a range of student models.
Similarly proceeding from creating examples, ~\cite{sahu2023promptmix} proposes PromptMix, which involves a two-step method based on prompting to create labeled examples for text classification. In PromptMix, the borderline examples can enhance the knowledge transfer from teacher models like GPT-3.5 to student models.
In contrast to the traditional unidirectional knowledge distillation, Lion~\citep{jiang2023lion} introduces an adversarial distillation framework, which encourages the teacher model to identify "hard" instructions and subsequently generate new "hard" instructions for the student model, resulting in a dynamic three-step adversarial cycle.

Black-box distillation is also identified as a promising tool to transfer the power of chain-of-thought (CoT) prompting from larger models to smaller ones.
~\cite{fu2023specializing} observe a trade-off in language models between their diverse capabilities, and focus on moving the teacher model's capability from general abilities towards enhancing the student model's proficiency in the targeted mathematical CoT.
SCOTT~\citep{wang2023scott} uses contrastive decoding for better rationale supervision and a counterfactual reasoning objective for faithful distillation, resulting in more faithful CoT rationales.
Distilling step-by-step~\citep{hsieh2023distilling} introduces a novel training method for smaller models, surpassing LLMs with less data. It uses LLM rationales as extra training material in a multi-task framework, cutting down data needs versus standard fine-tuning and distillation.
Similarly, ~\cite{li2023symbolic} propose symbolic CoT distillation, where they obtain CoT rationales for unlabeled dataset instances from the teacher model and then train the student model to forecast both the rationale and the label based on these instances.
To promote complex, multi-step reasoning within a dialogue context, i.e., dialogue CoT reasoning, ~\cite{chae2023dialogue} utilize LLMs as inconsistent teachers and strategically distill valuable and logical rationales through alignment filters.

\subsection{Factorization}
The use of low-rank matrix decomposition~\citep{kishore2017literature} as a technique for compressing Deep Neural Networks (DNNs) represents a straightforward yet effective approach, garnering considerable attention within both scientific computing and machine learning domains. In recent years, the challenge of efficiently compressing and accelerating large-scale neural networks via low-rank methods has become a focal point of research. This has led to significant advancements in developing and refining low-rank factorization strategies tailored for DNNs~\citep{schotthofer2022low}. 

Activation-aware Singular Value Decomposition (ASVD)~\citep{yuan2023asvd} is the first work using factorization techniques to compress LLM. 
ASVD effectively manages activation outliers by adjusting the weight matrix based on the activation distribution, improving decomposition accuracy and efficiency. ASVD also addresses the varying sensitivity of different LLM layers to decomposition, with an iterative calibration process for optimal layer-specific decomposition.
Concurrently, LAyer-SElective Rank reduction (LASER)~\citep{sharma2023truth} demonstrates the surprising result that it is often possible to significantly improve the performance of LLMs by selectively removing higher-order components1 of their weight matrices. 
Apart from targeting the LLMs' weights, TensorGPT~\citep{xu2023tensorgpt}, in which the embedding layer of LLMs is compressed through Tensor-Train Decomposition (TTD)~\citep{oseledets2011tensor} in order to store large embeddings in a low-rank tensor format, with much fewer parameters. 

%% file: sections/4_decoding.tex
\section{Algorithmic Methods for Fast Decoding} 
\label{sec:fast-decoding}

\begin{figure*}[h] 
    \includegraphics[width=\textwidth]{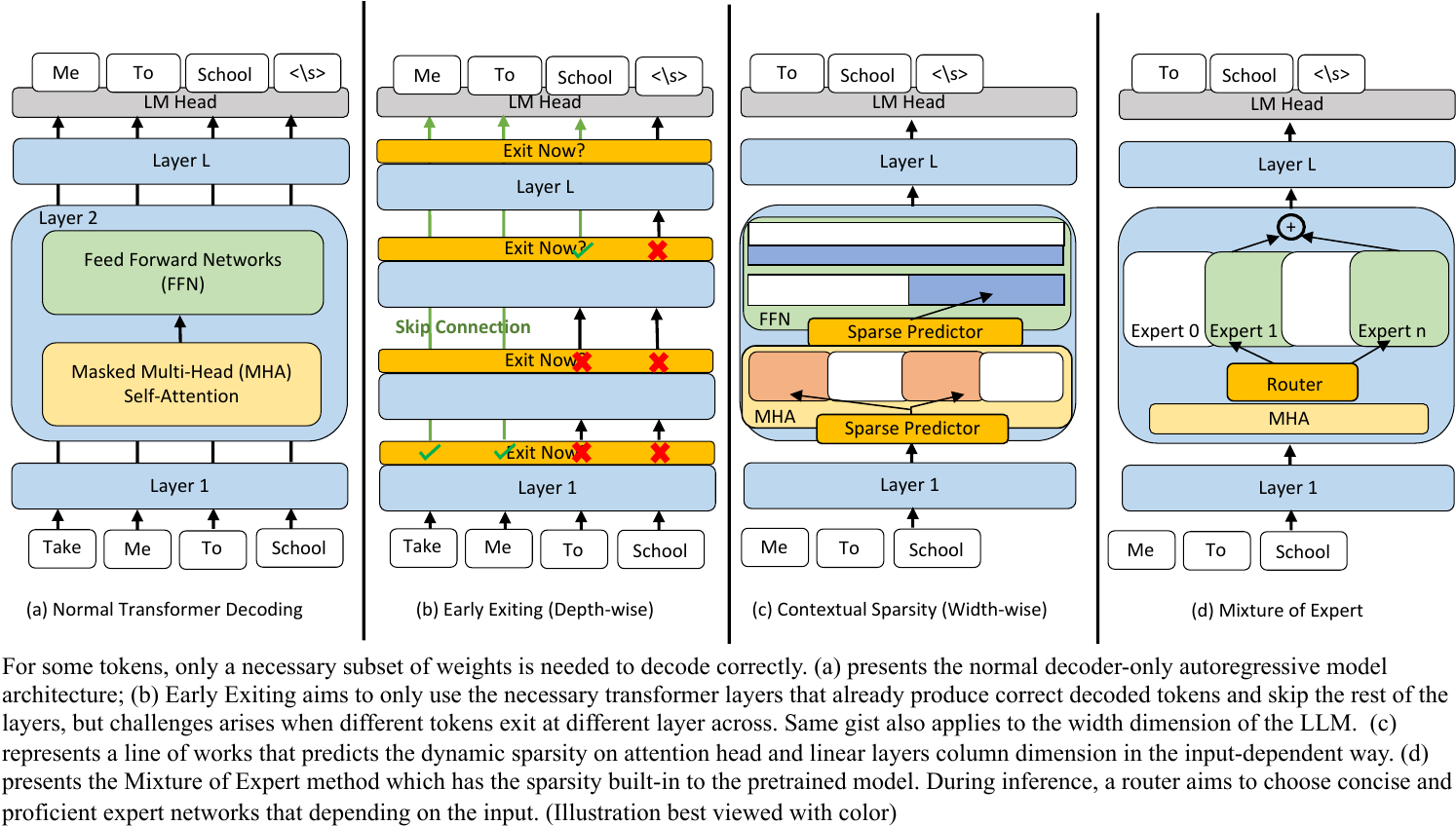} 
    \caption{Illustration of Input-Dependent Dynamic Network Technique} 
    \label{fig:early_exiting} 
\end{figure*} 

LLMs have achieved astonishing performance in various text generation tasks. They typically contain the decoder stage that generates tokens one after another, following an autoregressive relation with all preceding tokens. 
During the decoding of every token, decoder weights have to be repeatedly loaded into memory. As the parameter size of LLM becomes colossal, the decoding process becomes heavily memory-bound \citep{dejong2023fido} and experiences low hardware utilization, leading to severely long latency \citep{kim2023stack}. 
This is particularly problematic in real-world applications like ChatBot, where quick and even real-time responses are crucial. Therefore, there is a strong need to optimize the decoding process to improve performance in such applications.

This section focuses on the discussion of prior efforts to reduce the LLM inference cost from an algorithm perspective. 
Specifically, this section intends to develop the discussion from two directions: 
\begin{itemize}
    \item In section \ref{fixedtoken}, for every single token decoded (fixed \#tokens decoded), how to utilize the minimum number of parameters of the LLM. 
    \item In section \ref{fixedparameter}, for every single forward propagation of the LLM (fixed \#parameters used) how to decode the maximum number of tokens. 

\end{itemize} 

\subsection{Minimum Parameter Used Per Token Decoded} 
\label{fixedtoken} 

Interestingly, \cite{simoulin-crabbe-2021-many} has shown that although language models tend to have a huge number of parameters, not all parameters are needed to generate the accurate tokens. LLM inference latency can be reduced by selecting only a subset of necessary parameters to use (load) per input token and still preserving the decoded token's accuracy. In this section, we look at input-dependent dynamic weights dropping scheme for LLM from three different perspectives: \ref{earlyexit} looks at early exiting, or dynamically choosing weights in the layer, depth, dimension; \ref{contextualsparsity} introduces methods that dynamically detects sparsity in the width dimension of the LLM, pruning out heads and MLP columns; \ref{mioe} present Mixture-of-Experts (MoE) which pretrains a sparse model and chooses the correct experts for the different input during runtime.

\subsubsection{Early Exiting} \label{earlyexit}

Early exiting (or layer skipping) has been a well-explored idea in various network architectures, particularly for the encoder-only models~\citep{baierreinio2020node, hou2020dynabert, li2021cascadebert, liu2020fastbert, liu-etal-2022-towards-efficient, schwartz-etal-2020-right, stickland2019bert, xin2020deebert, zhou2020bert, zhu-2021-leebert, schuster2021consistent}.
Early exiting for decoder architecture requires consistency and quality retaining on the sequence level where each token depends on the previous tokens, which are considerations lacking in the previous abundant encoder-only early-exiting literature. 
The decoder contains layers of identical structure. Benefiting from this trait, the output hidden states of every layer can be used to pass in the LM Head to get a probability distribution prediction of the next token decoded. 
\cite{geva2022transformer} and \cite{simoulin-crabbe-2021-many} observe that for some tokens, the hidden states saturate during the intermediate layers. In other words, early exiting in the middle would, for some tokens, output the correct top-1 prediction as running through the full model. This observation lays the basis for the success of decoder early exiting methods. 

\cite{elbayad2020depthadaptive} conducts an early effort for efficient machine translation tasks to use early exiting on the decoder architecture. 
It proposes a general approach to follow. 
Shown in Figure~\ref{fig:early_exiting} (b), during the forward propagation, after every layer, there is an internal confidence function, usually a fixed metric or an MLP with a small number of layers, that computes a confidence score based on the hidden states on how likely it is to saturate at the current layer. The score is used to decide whether to exit through some carefully designed criteria. 
The LM Head is then used to output the next token-predicted probability distribution. 
Due to the high similarity of the newer follow-up works, we extend the discussion by looking at the key challenges of designing early exiting schemes for language models, where they introduce different novel techniques. 

\textbf{Modeling the Confidence of Saturation}. CALM \citep{schuster2022confident} studies three different ways to output the confidence score to exit: the softmax response, or the difference between the top two values after the softmax; the saturation of hidden states, or the cosine similarity between the current layer's hidden states with the last layer; the output of a linear classifier inserted to every layer. 
The linear classifier is trained by simply using a cross-entropy loss to align MLP output when inputting the hidden states with whether the top-1 token decoded as exiting on the current layer matches the top-1 token decoded of the full model. 
The experiments presented suggest that despite not being the most accurate predictor, the classifier method reaches the optimal trade-off between additional FLOPs overhead with prediction accuracy on score generating. 
Building up from CALM, \citep{bae2023fast} observed that when consistently exiting from shallow layers will result in an abnormally long length. Also, the confidence score computation on every layer injects high overhead and diminishes the benefit of early exiting. Therefore, it proposes to only have two choices for early exiting: either exit from the so-called "shallow module" or a group of shallow layers, or go all the way to the full model, or "deep module", drastically reducing the number of classifiers needed inside the model. Such design enables it to achieve more speedup than CALM, reaching 2x for certain tasks. 
On the other hand, ConsistentEE \citet{zeng2023consistentee} proposes a different method to predict when to exit. It uses an RL policy network that is iteratively trained with the per-layer output classifier head. The policy networks are trained with the goal of balancing the optimization of both efficiency (the early layer receives rewards) and accuracy (the reward function has a term that is the early exit output CE loss). 

\textbf{Early Exit Criteria}. CALM \cite{schuster2022confident} proposes a distribution-free calibration technique that uses the fixed sequence testing procedure (Family-wise Error Rate procedure) to output the suitable threshold. The threshold is exponentially decreasing to allow more aggressive exiting for tokens later in the sequence. \cite{bae2023fast}, on the other hand, observes that the pattern of confidence criteria resembles a beta distribution and uses the on-the-fly data to update a beta distribution model through MLE and use such probability model to guide its decision.
\cite{zeng2023consistentee} bypasses this issue by letting the policy network directly output the exit decision. 

\textbf{Hidden States Propagation}.
Hidden states of the skipped layers can pose a technical challenge. As shown in the \ref{fig:early_exiting} (b), the token position at "school" exits later than previous tokens. However, the last self-attention layer doesn't have the previous key-value pairs of the previous early exited tokens. \cite{elbayad2020depthadaptive} and \cite{schuster2022confident} proposes the "hidden states propagation" technique. For example, the hidden states of token "Max" at the exited layer $l_1$ are stored. When the later token "school" reaches deeper layer $l_2$, the hidden state for "Max" is copied for all layers between $l_1$ and $l_2$, and the key-value pairs are then computed on the copied hidden states. Basically, to approximate the deep layer's hidden state with the ones from the early layer. Later works \cite{bae2023fast} and \cite{ding2023efficiency} found that state propagation leads to performance degradation. Since LLM inferences are dominated mostly by memory loading, computation is relatively "free". These two methods proposed to recompute the later hidden states directly on the fly. \cite{chen2023eellm} proposes to run the full large model in parallel to the early exit stream to efficiently parallel the computation of the missing kv cache. \cite{din2023jump} conducts a systematic study on using a linear network to jump across layers for transformer architecture and shows that linear layers can be added to effectively bridge the performance gap between directly copying and computing the hidden states with low memory and compute cost. 
SkipDecode \cite{delcorro2023skipdecode} chooses an aggressive approach to prioritize the speedup and relax the performance preservation goal. By utilizing the observation that a token coming later in the same sequence on average requires less number of layers to decode the correct tokens, it completely bypasses the need for state propagation by forcing the maximum layer used to be monotonically decreasing for deeper positions. Besides, SkipDecode also introduces fixed exit points to optimize for batched early exit. 

\textbf{Output Classifier Training}. When exiting from intermediate layers, the intermediate hidden states need to go through an output classifier head to output prediction of the next token probability distribution. The output classifier can either be shared as shown in Figure~\ref{fig:early_exiting} or per-layer independent. These classifiers are usually trained to better adapt to the early exiting pattern. \cite{elbayad2020depthadaptive} proposed to have an average CE loss of all layers to be the training loss of the classifier. On the other hand, \cite{schuster2022confident} uses a weighted average where weights increase as the layer number increases, assigning more contribution to deeper layers. \cite{bae2023fast} introduces a dynamic knowledge distillation loss which dynamically assigns the "shallow module" a suitable hidden state from the "deep module". Both \cite{rotem2023finding} and \cite{ji-etal-2023-early} find a "conflicting gradient" issue when joint training with the same loss across all models: \cite{rotem2023finding} detects the gradient conflict between early and later layers of language models, while \cite{ji-etal-2023-early} spots the "orthogonal gradient" between the objective to improve semantic awareness and the objective to improve early exciting decision. Both methods propose adding an additional block of parameters and iterative training to alleviate the issue. 
Besides the above-mentioned perspectives, \cite{chen2023eellm} studies system-level optimization techniques to efficiently run LLM early exit under the 3D parallelism setting. 


\begin{figure*}[ht] 
    \includegraphics[width=\textwidth]{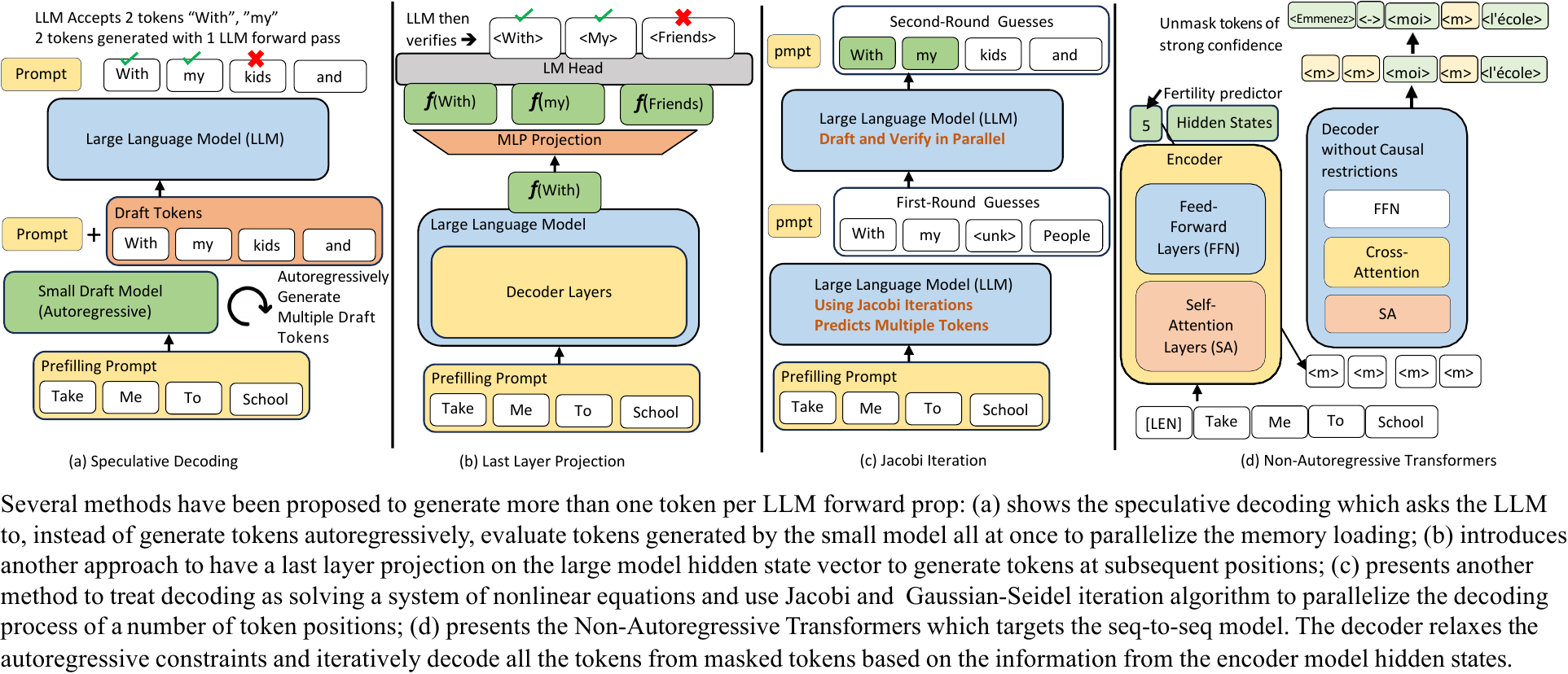} 
    \caption{Illustration of the Parallel Decoding Methods} 
    \label{fig:specudec} 
\end{figure*}

\subsubsection{Contextual Sparsity} \label{contextualsparsity} 

While early exiting aims to select parameters on the depth dimension, some techniques have also been proposed to exploit the dynamic sparsity on the width dimension. Deja Vu \cite{liu2023deja} conducts a comprehensive study on dynamic sparsity on the LLM width dimension. The paper reveals that contextual sparsity can go up as high as 80\%, meaning that the majority of the weights can be left out while still preserving the original model performance. However, the chosen weights are dynamic and different for different input tokens. The paper formulates this problem as a nearly neighbor search problem that for a given hidden state from the embedding layer of previous layers, how to find the attention heads and the MLP columns that are the most similar to these tokens. To save compute, the paper proposes to train a small MLP network as the Sparse Predictor in front of the Multi-Head Attention (MHA) and the Feed-Forward Networks (FFN) of the LLM, shown in Figure~\ref{fig:early_exiting} (c). By using only a subset of weights and reducing the memory IO overhead, Deja Vu manages to achieve over 2x speedup of LLM inference. Building on Deja Vu, PowerInfer (\cite{song2023powerinfer}) brings the contextual sparsity finding to the LLM inference across heterogeneous devices (CPUs and GPUs). PowerInfer discovers a substantial portion of weights are heavily used and activated in the input-independent setting, thus stored on the GPU memory, while others are on the CPU memory. 
Then, to specifically find the weights to use for a given input token, it trains a smaller sparse prediction than Deja Vu. To make the sparse predictor, it initializes the sparse predictor to have a dynamic structure and iteratively trains and modifies the sparse predictor. 
To better do inference of the model deployed on the mixed CPU and GPU environment, it introduces a novel memory placement scheme and implements a vector-based sparse computation library. 
Concurrently, MatFormer (\cite{devvrit2023matformer}) studies the problem of LLM deployment on various heterogenous devices of different hardware capabilities. They added dynamic structure only on the FFN, which occupies 60\% of total weights. The model is specially trained so that during inference, based on the target hardware properties, MLP layers are sampled on the row dimension to give a model of various sizes with reasonable performance. 
To diversify the model size selection, it imposes a Mix’n’Match method to choose different settings for different layers, so combined would give a more variable model size. 

\subsubsection{Mixture-of-Expert Models} \label{mioe} 
Language Models, especially transformer architecture, exhibit strong power-law scaling (\cite{kaplan2020scaling}, \cite{hoffmann2022training}) of performance when the training dataset is scaled up. On the other hand, though bringing strong performance gain, the large parameter count makes training and inference of the model inefficient. The mixture of expert (MoE) technique is a well-studied topic (\cite{6215056}) that effectively decouples the parameter count of the model and the computation FLOPs required by the model training and inference, thus bringing huge gains of efficiency under certain conditions. Further, MoE is shown to effectively scale up the language model size and increase its performance without the concern of increasing compute during inference (\cite{lepikhin2020gshard}, \cite{fedus2021switch}). Shown in Figure \ref{fig:early_exiting} (d), an expert network is inserted into the transformer architecture to replace the FFN layers. Also, a gating function is introduced between the Multi-Head Attention and the expert network which aims to select the best-fit expert or experts for the given input token. For in-depth analysis and discussion about the MoE scaling generalization, routing algorithms, training techniques, etc., we refer the readers to the survey on Sparse Expert Models (\cite{fedus2022review}). 
\textit{Although both rely on the input token to determine sparse structure, We deliberately separate MoE and the contextual sparsity techniques because the latter operates on pre-trained dense language models and exploits the sparsity from the dense neural networks, while the prior trains a sparse model from the beginning.} 
More recently, MoE techniques have achieved substantial success. Sparse Mixer (\cite{lee2022sparse}) brings 89\% and 98\% speedup in both training and inference to BERT (\cite{devlin2019bert}) models. \cite{du2022glam} uses only 49\% FLOPs but beats GPT-3 (\cite{brown2020language}) in performance. ST-MoE (\cite{zoph2022st}) brings MoE to the encoder-decoder models, even becoming the state-of-the-art model for many reasoning and generation tasks. ST-MoE, using 
20x and 40x fewer FLOPs in training and inference, beats 540B PaLM (\cite{chowdhery2022palm}) in performance. Mixtral 8x7B (\cite{jiang2024mixtral}), while only actively using 13B parameters during inference, performs on par with Llama2-70B models (\cite{touvron2023llama2}) across various evaluation benchmarks. 

Besides, various attempts have been made to optimize MoE model inference. \cite{kossmann2022optimizing} builds an efficient compiler library RECOMPILE for MoE models that introduce dynamic recompiling and optimization according to varying inference batch sizes. \cite{rajbhandari2022deepspeed} extends the ZeRO distributed inference method to MoE models. \cite{jawahar2023automoe} conducts Neural Architecture Search (NAS) on the expert network architecture. \cite{yi2023edgemoe} deploys large MoE language models on the edge devices. It optimizes the deployment around the finding that some neurons are much more heavily used in the MoE models than others. 

\subsubsection{Roofline Model Analysis for Dynamic Parameter Reducing}


The Minimum Parameter Used Per Token Decoded methods simultaneously decrease computational and memory access overhead. From the viewpoint of roofline model, these methods result in small changes to the arithmetic intensity of each operator and the type of bound.

For Early Exiting or Layer Skipping methods, entire Transformer layers are skipped, leading to a proportional reduction in overall computation, memory access, and inference time. In other words, the inference time decreases proportionally to the number of layers skipped in the network.
However, for methods like Contextual Sparsity and Mixture of Experts, the arithmetic intensity varies across different operations. Consequently, dynamically choose to activate these layers leads to varying reductions in computation and memory access, resulting in different impacts on the overall inference time.

\subsection{Maximum Tokens Decoded Per LLM Forward Propagation} 
\label{fixedparameter}

Another angle to reduce the latency of LLM inference is to relax the LLM from the limitation of autoregressive decoding and have more than one token decoded per one LLM forward propagation. We look at two ways to achieve it: \ref{specdec} presents the speculative decoding method which introduces a computationally efficient draft model to propose candidates for the next few token positions, while the LLM is used to evaluate the draft model's proposed draft tokens, instead of generating next tokens. 
On the other hand, \ref{pd} presents works that enable the LLM to directly decode multiple tokens from a single forward propagation. Due to some methods combining the benefits from both directions and lying in the middle, we manually add a distinction just for the sense of nomenclature that speculative decoding methods here all have the draft model to be in the transformer architecture. 

\subsubsection{Speculative Decoding} \label{specdec} 

Due to the demanding memory loading challenges and autoregressive properties, LLMs are inefficient in inference. However, models that are much smaller in size are shown (\cite{kim2023speculative}) to have the ability to decode the correct sequences as the LLM, as long as some key tokens in the sequence of the small model generation are corrected. Then, shown in Figure \ref{fig:specudec} (a), when the small model is asked to infer (speculate) and output a sequence of draft tokens, memory loading of model weights is less of a problem, resulting in much higher utilization in hardware computation units. To ensure the quality of the text generated by the small model, the LLM can "periodically" evaluate and correct tokens from the small model's draft. Then, although the large model needs to sometimes evaluate the wrong draft tokens, potentially leading to larger FLOPs spent than LLM autoregressive decoding, the memory loading of weights is parallelized on the token dimension and drastically reduces the memory IO overhead. Since the LLM inference is memory bottlenecked, the speculative decoding will potentially reduce the LLM inference latency greatly. 

\textbf{LLM Distribution Preserving} During early exploration of this idea, two different paths emerged concurrently. \cite{kim2023speculative} proposed to have the small model speculate and generate draft tokens until the token decoded confidence falls below a threshold. Then, the small model "fallback" to the large model to evaluate the draft tokens generated and hand over to the small model. Some of the tokens are rejected, so the large model asks the small model to "roll back" these wrong tokens and resume speculating. In the paper's setting, all decoding is "greedy". The paper show that the large and small model pair can generate text with quality on par with the original large model autoregressive generated text. However, \cite{leviathan2023fast} and \cite{chen2023accelerating}, upon the small model speculate paradigm, points out a technique of resampling that at the position where the LLM rejects the small model's prediction that provably enables the large and the small model predictions to be in the same probability distribution as the large model's autoregressive generation. 
The following techniques generally follow the paradigm of speculating then evaluating and resampling to preserve the LLM autoregressive decoding quality while enabling speedup. 

\textbf{Building a Tree of Draft Tokens} Since the LLM generates in the autoregressive order, every token is dependent on all previous tokens generated, and the length of the accepted tokens in the small model's draft is usually modest and bounded. It is exponentially more difficult to speculate on tokens more distant in the future. For example, if the small model is asked to output the length m draft sequence, and the LLM accepts n, n $<$ m, the (m - n) tokens are automatically discarded. 
Thus, the speedup ratio of speculative decoding is modest, since every LLM forward leads to only a limited number of tokens being decoded. 
There are two ways to improve the speedup of speculative decoding. First, \cite{sun2023spectr}, \cite{miao2023specinfer}, and \cite{xu2023llmcad} all proposed to boost the draft on the batch size direction, or letting the small model sample multiple plausible draft sequences for the LLM to evaluate in parallel. Specifically, \cite{sun2023spectr} proposes a way and theoretical guarantees for the LLMs to batch verify and resample from the multiple small model drafts so that the LLM distribution is preserved and no loss of generation quality is incurred. The paper first connects speculative decoding to the broader problem of discrete optimal transport. The small model is asked to sample multiple draft sequences using topk sampling. Based on the properties of the discrete optimal transport, finding the optimal method to evaluate and resample becomes finding the optimal transport path. On the other hand, besides from maintaining the speculative decoding consistency of draft trees, \cite{miao2023specinfer} constructs the token tree not based on the top predictions from the small draft model, but based on multiple diversely trained small draft models, each running in parallel and output diverse but powerful draft sequences. The paper proposes a novel draft token tree construction algorithm that builds a tree of candidate tokens based on the diverse draft sequences through predefined expanding and merging schemes. Then, the large model is asked to parallel verify the constructed tree using a carefully designed tree attention to maximize the reuse of the key-value cache and maintain a tree-based causal mask. \cite{xu2023llmcad} innovatively applies the benefit of speculative decoding to edge devices. The paper builds an LLM serving engine for the edge, where a smaller draft LLM is sitting consistently in memory, while a larger robust LLM is occasionally loaded in memory to do verification. To boost the acceptance rate from the large LLM, it also constructs a tree using topk tokens. To cater to the edge hardware characteristics, it implements a tree-based parallel verification decoder equipped with masking and a customized large-small LLM computation pipeline to avoid memory contention. 

\textbf{Knowledge Distillation and Self-Speculative Decoding} Another way to improve the acceptance rate is to improve the small draft model's ability to align with the LLM's generation distribution, which can be done through finetuning the small models on corpus generated by the large models with knowledge distillation. 
\cite{zhou2023distillspec} establishes a mathematical connection between the acceptance rate and natural divergence between the small model and the LLM: minimizing the divergence is maximizing the acceptance rate. The paper also studies a range of different knowledge distillation losses and shows that adding knowledge distillation brings consistent 10-45\% improvement in latency speedup. However, the paper generally finds that the optimal knowledge distillation loss choices vary model by model and should be tuned as a hyperparameter. \cite{liu2023online} also shows that knowledge distillation boosts the small model training. Besides, the paper brings speculative decoding to the cloud online learning settings. LLM inference is memory-bottlenecked, which means that there is always a surplus in computation resources. 
The compute can be used to train a draft model continously on server, which brings two benefits: Continuously training with knowledge distillation boosts its acceptance rate and, thus, reduces the LLM inference latency; 2) serving input is constantly shifting in domains, and continuous training helps the draft models maintain the strong performance in different domains. 
\cite{zhang2023draft} avoids storing a separate draft model by selectively sampling a smaller draft model from the large model itself. Before deployment, the paper utilizes a Bayesian optimization method to search for a draft model by skipping intermediate layers within the pretrained large model. Besides, it proposes an adaptive threshold selection technique tailored for the decoding of the draft model sampled from the large models. 

\subsubsection{Parallel Decoding} \label{pd} 

Alternatively, abundant works have been proposed to enable the large model to directly perform parallel decoding without the help of a small transformer model. 

\textbf{Simultaneously Predicting Multiple Future Tokens} A wide variety of works are exploring the subject of enabling multiple token predictions directly from one forward pass of the Large Language Model. \cite{stern2018blockwise} pioneers the design of inserting a linear projecting layer between the last hidden states output and the input of the language modeling head to enable multiple future tokens to be projected solely based on the current token's last hidden states as input. Evaluation is subsequently made by the LLM to decide whether to accept or reject these projected tokens. The proposed technique focuses on the sequence-to-sequence models that have the decoder structure. 
More recently, \cite{cai2024medusa} extends the previous work to the decoder-only language models as shown in Figure \ref{fig:specudec} (b). Besides the last layer projection, to further improve the decoded acceptance rate, the paper proposes to add a tree-based decoding structure and the associate attention mask design to propose multiple drafts simultaneously for the large model to evaluate. 
Besides, concurrently \cite{monea2023pass} proposes to add several dummy tokens at the end of the input sequence are called "lookahead embeddings" in work. 
During the forward pass of each layer, the information of previous prompt tokens and already decoded tokens can be used to parallel decode several consecutive future tokens. 
To enable this design, the work trains a separate embedding layer that specifically serves these lookahead embeddings. \cite{li2024eagle} also aims to do parallel decoding with LLM evaluation. Like previous works, it also adds a lightweight structure FeatExtrapolator. Differently, the structure takes both the previous token's last layer hidden states and the actual decoded token embedding as input and output the hidden states prediction of the next layer. The LM head of the LLM is used, and several tokens are sampled, which are then used to build a decoding tree for the LLM to evaluate in parallel. 

\textbf{Retrieval of Frequent N-grams} Besides directly using the LLM to output several following tokens, some works use the frequently appeared n-grams in natural language to enable multiple future tokens to be generated within one forward pass of the large model. LLMA (\cite{yang2023inference}) first observes that the generation tasks tend to ask the LLM to repeat tokens that appeared in the previous contexts. Based on this information, the paper set out to use the decoded tokens and the prompt to do prefix matching with a set of reference documents so that if a repetition occurs, tokens that are repeated can be directly copied to the current place. Then, an LLM will evaluate these found candidate tokens from the previous context to decide whether to use them. \cite{he2023rest} further extends LLMA and proposes to first construct a database of common phrases based on the LLM pretrained or finetuned dataset and corpus. Then, during decoding, the previous context prompts or tokens are used as the query to be used to retrieve into the constructed database. The candidates retrieved are organized into a prefix tree structure or a trie, which the LLM can then evaluate efficiently. \cite{lan2023copy} similarly follows to use the retrieval methods to speed up inference. In contrast, it adds an extra attention layer at the end of the LLM to use the current context represented by the hidden states of the current token as the query to attend to relevant phrases retrieved from the documents of reference and select top phrases based on the attention scores. 

\textbf{Hierarchical Structure In Language} Hierarchical Structure exists in language. For writing a long piece of article, the usual approach is to first write out the general outline of the paper, as in the format of bulletin points. Then, for every bulletin point, arguments can be extended to encapsulate the full intent of the bulletin point. Based on the observation that arguments for different bulletin points are relatively independent in semantics, some methods are proposed to parallelize the generation process for different bulletin points. Skeleton-of-Thoughts (\cite{ning2023skeletonofthought}) proposed to first ask the LLM to generate concise bulletin points for an article, and then collect these bulletin points on the batch axis and feed them into the LLM again as a prompt to ask the LLM to expand the arguments for each bulletin points in parallel. The achieved speedup is approximately 2x, but with the caveat that the method cannot easily generalize to all text generation tasks. More recently, APAR (\cite{liu2024apar}) extends upon this direction. The paper adds specific soft tokens that explicitly inform the LLM of the hierarchical information during the generation. The LLM is further instruct-tuned to incorporate the added special tokens, and the generation is boosted with the Medusa (\cite{cai2024medusa}) technique to achieve 4x speedup on text generation with the hierarchical structure. 

\textbf{Jacobi and Gaussian-Seidel Iterative Algorithms} \cite{pmlr-v139-song21a} pioneers the study of using parallelizable methods to approximate the results from iterative and sequential inferences of fully connected networks or CNNs. Though seemingly in-viable, the paper finds that neural networks can tolerate numerical approximation errors and the data patterns that neural networks learn expose parallel structures to some extent, which makes it possible in some scenarios to parallelize the sequential inference of neural networks. 
Jacobi and Gaussian-Seidel Algorithms were previously proposed to solve a system of non-linear equations (\cite{ortega2000iterative}) and are shown to effectively parallelize the sequential neural network inference. \cite{santilli2023accelerating} extends the Jacobi and Gaussian-Seidel algorithms to parallelize the autoregressive decoding in the Machine Translation tasks. Specifically, this work is built on top of the previous Non-Autoregressive Transformers architecture (which we will cover later in the chapter) to enhance the parallel decoding with GS-Jacobi algorithms. The parallel decoding process stops when a [EOS] token is found in the decoded text. Concurrently, Lookahead decoding (\cite{fu2023lookahead}) shown in Figure \ref{fig:specudec} (c) extends this method to parallelize the LLM generation of subsequent tokens. Besides using the vanilla Jacobi iterative algorithm, it also boosts its speed with a retrieval-based algorithm to reuse the previously seen n-grams. In addition, it parallelizes the lookahead step and LLM verification step by introducing a carefully designed attention mask to the original LLM model to further improve decoding efficiency. 

\textbf{Non-Autoregressive Transformers} For Machine Translation tasks that require autoregressive decoding of the sequence-to-sequence model, Non-Autoregressive Transformers (NAT) has been proposed to iteratively decode all of the output tokens together, as shown in Figure \ref{fig:specudec} (d). NAT has been relatively well-explored (\cite{gu2017non}, \cite{wang2019non}, \cite{li2019hint}, \cite{sun2019fast}, \cite{wei2019imitation}, \cite{shao2020minimizing}, \cite{lee2018deterministic}, \cite{ghazvininejad2019maskpredict}, \cite{guo-etal-2020-jointly}, \cite{gu2020fully}, \cite{savinov2021step}), and we point the readers to the following survey paper that covers specifically NAT models \cite{xiao2023survey} for an in-depth review and analysis on the subject. 
Coarsely, the speedup of text decoding comes from making a single forward pass of the decoder output more than one token. The input sequence is first fed into the encoder, which outputs the hidden states that extract the input semantics. The output hidden states of the encoder are then used as the condition for the decoder pass. To speed up the text generation, the decoder side relaxes the autoregressive constraints and takes a sequence full of dummy tokens [pad] as the input to start the iterative parallel decoding process. During each iteration, based on the condition set by the encoder output hidden states, some tokens can be confidently predicted, which are unmasked. 
The sequence is mixed with unmasked decoded tokens and the remaining masked tokens are fed to the decoder again until every token is decoded. 
The length of the sequence fed into the decoder, or fertility, is usually learned either inside the encoder as a special [CLS] token or by a specialized fertility predictor between the encoder and the decoder. More recently, \cite{savinov2021step} treats the decoder as a diffusion model and trains it to denoise the noisy initial sequence based on the conditions given. However, because of the requirement to use encoder hidden states as the condition for parallel decoding, NAT methods face natural difficulties in extending directly to decoder-only architectures. 


%% file: sections/5_system.tex
\section{Compiler/System Optimization} 
\label{sec:system_optimization}

After model compression and algorithm optimization for LLMs, the next step is to compile and deploy them on hardware devices. To ensure efficient inference of LLMs, there are various compiler optimizations that can be employed.
Moreover, due to the increasing scale of LLMs, multiple hardware devices may be required for deployment and execution, forming a complex inference infrastructure system. As a result, system-level optimization for efficient inference has become a hot topic.
In this section, we will explore some widely used compiler optimization and system optimization techniques. 
These include operator fusion, memory management, workload offloading, and parallel serving.


\subsection{Operator Fusion}

\begin{figure}[t]
    \centering
    \includegraphics[width=0.95\linewidth]{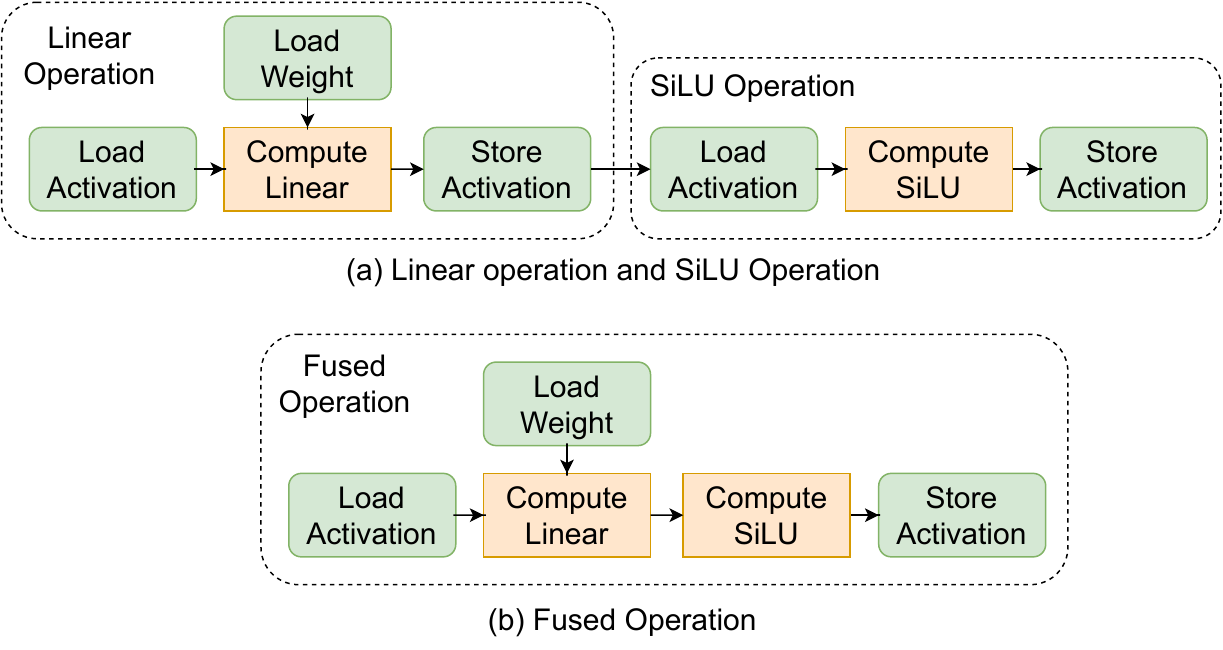}
    \caption{Demonstration of operator fusion for a linear operator followed by a SiLU operator.}
    \label{fig:operator_fusion}
\end{figure}

\begin{figure}[t]
    \centering
    \includegraphics[width=0.95\linewidth]{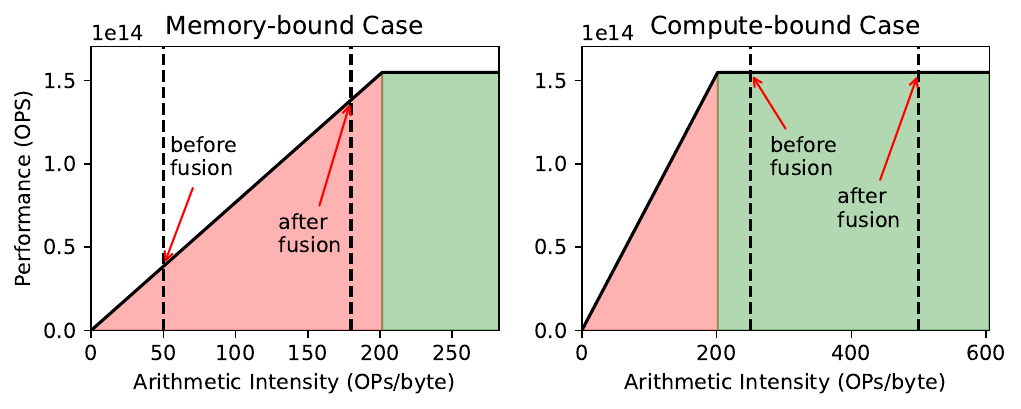}
    \caption{Demonstration of the memory-bound case and compute-bound case for operator fusion.}
    \label{fig:roofline_operator_fusion}
\end{figure}


\begin{figure}[t]
    \centering
    \includegraphics[width=0.95\linewidth]{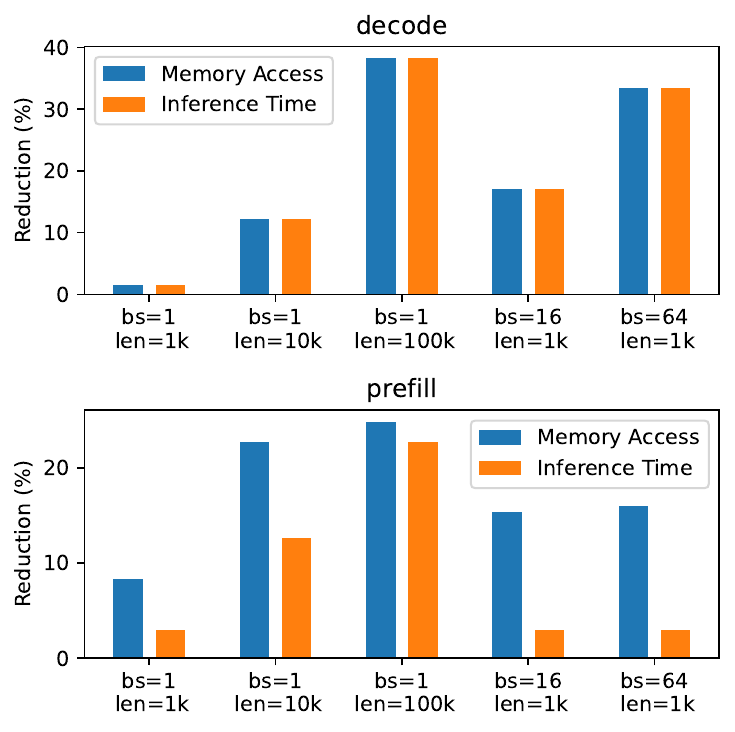}
    \caption{The memory access reduction and inference time reduction for FlashAttention on Nvidia A6000.}
    \label{fig:flashattention_reduction}
\end{figure}

Operator fusion is an important compile-time optimization technique in deep learning frameworks to improve computational efficiency.
It combines together multiple operators or layers that are directly connected in the computation graph. 
This eliminates redundant data movement and intermediate representations.
For example, a linear operator followed by a SiLU operator can be fused together into a single operator. 
As shown in Figure~\ref{fig:operator_fusion}, this avoids having to store and load the intermediate activations between each operator, reducing both the memory consumption and the memory access.
As shown in Figure~\ref{fig:roofline_operator_fusion}, the roofline model suggests that kernel fusion can increase arithmetic intensity and enhance inference performance in memory-bound areas.
However, when operators are already in a compute-bound area, memory fusion provides little benefit.

While operator fusion can provide significant performance benefits in many cases, it is not applicable to all operators. Operator fusion may not be possible or beneficial for certain operators:
(1) Operator fusion requires that the intermediate results of the fused operations are not needed elsewhere in the computation graph. If a subsequent operation depends on the output of an intermediate operation, fusion is not possible without introducing additional complexity or recomputation.
(2) Operator fusion can potentially increase the on-chip buffer requirements of the fused operation. If the available on-chip buffer is limited, it may not be feasible to fuse certain operations.
(3) Some frameworks or hardware architectures may have limitations or restrictions on which operations can be fused together, depending on their implementation details.


Some compilation tools, such as TVM~\citep{chen2018tvm}, are capable of identifying operators that can be fused together and replacing them with a fused operator. 
However, for LLMs, automatically detecting and fusing operators is both unnecessary and complex because LLMs have a fixed architecture. 
Instead, specific fusion patterns can be used to improve efficiency. 
For instance, the attention mechanism is an essential part of LLMs.
Automatically fusing attention mechanism can be a complex task for compilation tools.
FlashAttention~\citep{dao2022flashattention,dao2023flashattention2} and Flash-Decoding~\citep{dao2023flashdecoding} proposed fusing the matrix multiplications and softmax operator in self-attention into one operator.
This fusion technique eliminates the need to store and load the intermediate attention matrix, which can be very large when the sequence length or batchsize is large. 
As shown in Figure~\ref{fig:flashattention_reduction}, fusing them can significantly decrease the memory access and inference time.
We can observe that there are differences between the prefill stage and decode stage. 
In the decode stage, the memory access reduction is the same as inference time reduction. 
However, in the prefill stage, inference time reduction is lower than memory access reduction. 
This is because some operations in the prefill stage are compute-bound, so reducing memory access by operator fusion provides little benefit.

DeepSpeed-inference~\citep{aminabadi2022deepspeedinference} introduces a technique called Deep-Fusion.
It specifically fuses four main regions within a transformer layer: the QKV GeMM and input layer normalization; transposition and attention operations; post-attention layer normalization and intermediate GeMM; bias addition and residual addition.
xFormers~\citep{xFormers2022} offers various fused kernels that can enhance the performance of transformers. These include fused softmax, fused linear layer, fused layer norm, and fused SwiGLU. 
TensorRT-LLM~\citep{vaidya2023tensorrt_llm} is another framework that offers a wide range of high-performance fused kernels. It incorporates a powerful pattern-matching algorithm that can detect potential fusions in various LLMs. 

In addition to kernel fusion, we can enhance the performance of the LLM by further optimizing operators' implementation. 
For example, FlashDecoding++~\citep{hong2023flashdecoding++} proposes using asynchronized softmax and flat GEMM optimization with double buffering to improve efficiency.


\subsection{Memory Management and Workload Offloading}

When using an LLM to generate responses, the number of input and output tokens can change each time.
The length of the user's input prompt may vary, affecting the length of the sequence in the prefill phase. Additionally, the sequence length increases incrementally during the decode phase as tokens are generated.
This means that the shapes of the activations are not fixed like in a normal neural network. 
How to manage the memory efficiently as the tensor sizes change is a problem. 
PagedAttention~\citep{kwon2023efficient} efficiently handles the KV cache by dividing it into blocks.
The KV cache of each sequence is divided into blocks, with each block containing the keys and values for a fixed number of tokens.
To manage these blocks, a table is used to map the logical blocks of a sequence to the physical blocks in GPU memory.
This mapping is similar to how virtual memory works in a CPU's memory management system.

\begin{figure}[t]
    \centering
    \includegraphics[width=0.9\linewidth]{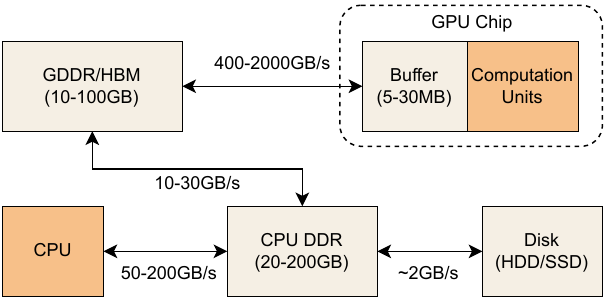}
    \caption{In typical computer architectures, the memory system consists of different types of memory spaces.}
    \label{fig:computer_arch}
\end{figure}
\begin{figure}[t]
    \centering
    \includegraphics[width=0.9\linewidth]{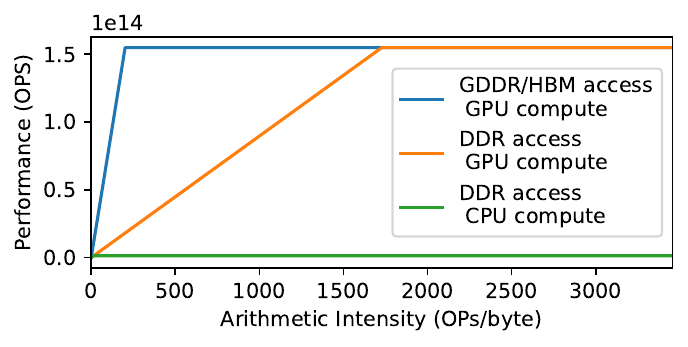}
    \caption{Roofline model for different offload settings.}
    \label{fig:offloading_roofline}
\end{figure}


When the GPU has limited memory capacity and the network is too large to fit, it may be necessary to employ workload offloading to store the network in alternative memory spaces.
As depicted in Figure~\ref{fig:computer_arch}, a computer system consists of various memory spaces, including CPU's DDR, GPU's GDDR/HBM, and hard disk. However, these different memory spaces have distinct access bandwidths.
Figure~\ref{fig:offloading_roofline} illustrates that when the data is offloaded to CPU's DDR and transferred to the GPU for computation when needed, it is better than performing the computation on the CPU. When the batch size is large enough, the arithmetic intensity increases significantly, allowing the GPU to fully utilize its computation capacity and achieve good results.
DeepSpeed-inference~\citep{aminabadi2022deepspeedinference} introduces ZeRO-Inference, which offloads the weights of large models to CPU memory.
This mechanism performs well with large batch sizes because the increased batch size increase the computation requirement and make the computation latency overlap the latency of fetching model weights, thereby improving overall efficiency.
Huggingface Accelerate~\citep{huggingface2022-accelerate} can also move certain modules to the CPU or disk if there is not enough GPU space to store the entire model.
FlexGen~\citep{sheng2023flexgen} provides a way to explore different ways of offloading computations considering constraints imposed by available hardware resources from the GPU, CPU, and disk. To find the best strategy in terms of throughput, FlexGen employs a linear programming-based search algorithm.
\citet{alizadeh2023llminflash} takes advantage of the larger capacity of flash memory compared to DRAM. It efficiently performs inference by storing model parameters in flash memory and transferring them to DRAM when needed.

\subsection{Parallel Serving}

\begin{figure}[t]
    \centering
    \includegraphics[width=0.9\linewidth]{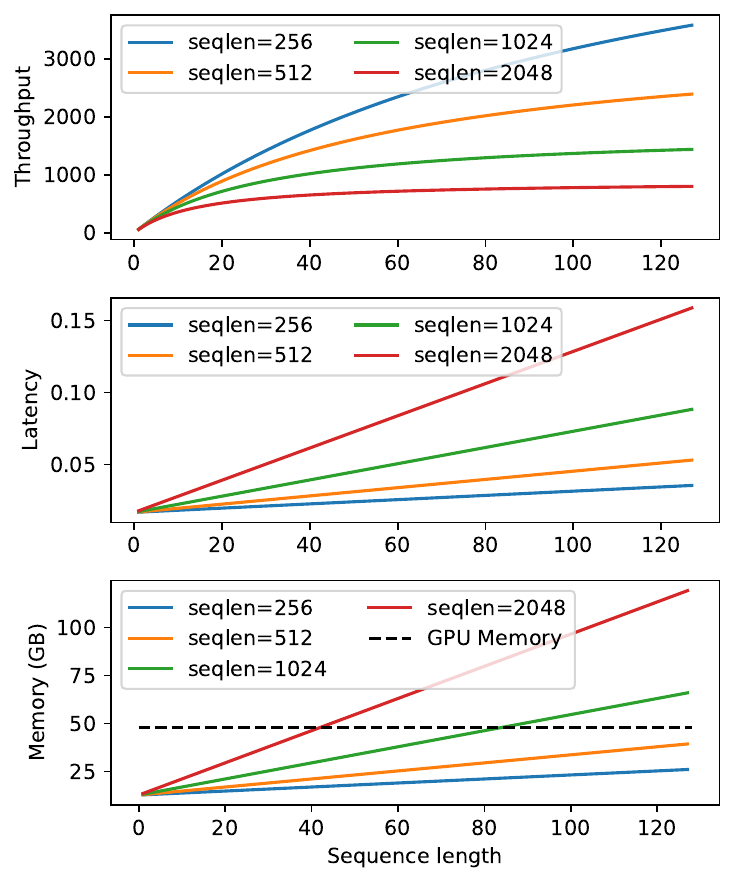}
    \caption{The parallel serving settings have an impact on the throughput, latency, and memory usage of the Nvidia A6000 GPU (Llama-2-13b).}
    \label{fig:parallel_serving}
\end{figure}

Parallel serving handles multiple user requests to a server at the same time. One goal is to respond to each request quickly. To achieve this, we need to reduce the time it takes to respond to each user, known as the response latency.
Another important factor to consider is throughput, which is the number of requests the server can process in a given time. By increasing the server's throughput capacity, we can serve more users simultaneously, leading to better overall system performance.
By increasing the server's throughput capacity, more users can be served simultaneously, resulting in improved system performance.
The serving system should be optimized to maximize throughput, while still ensuring that the response latency is within acceptable limits. 
Batching is a fundamental approach to improve throughput by processing multiple user requests together. Figure~\ref{fig:parallel_serving} shows that increasing the batch size during the decode stage significantly enhances throughput. 
However, increasing batch size can increase the response latency and memory consumption.

Several techniques have been proposed to optimize the batching method. For example, ORCA~\citep{yu2022orca} introduces continuous batching (also known as iterative or rolling batching) to combine inferences from different users. 
SARATHI \citep{agrawal2023sarathi} employs chunked-prefills and decode-maximal batching.
It combines prefill chunks and decode requests to create batches, which increases the arithmetic intensity and improves throughput.
Similarly, DeepSpeed-FastGen \citep{holmes2024deepspeedfastgen} and LightLLM \citep{modeltc2024lightllm} also employ a split and fuse technique.


%% file: sections/6_hardware.tex
\section{Hardware Optimization}
\label{sec:hardware-optimization}


Designing hardware to efficiently support inference for LLMs is a challenging task due to the varying arithmetic intensity\footnote{Arithmetic intensity refers to the ratio of arithmetic operations to memory access, which has been described in Roofline model (Section~\ref{sec:Rooflinemodel}).} under different inference stages and workload conditions. 
Specifically, the prefill stage usually leverages GEMM operators to process the batched tokens, which exhibits high arithmetic intensity. On the contrary, the decoding stage calculates output tokens one at a time, which necessitates the use of either GEMV operators or lean GEMM operators to process the attention and FFN layers. These operators are characterized by low arithmetic intensity.  

Furthermore, the arithmetic intensity can exhibit substantial variation depending on the batch sizes and sequence lengths. For instance, a large batch size could significantly alter the arithmetic intensity, and a long sequence length may increase the memory access overhead of  KV-cache reading in each decoding step. This variability introduces additional complexity into the hardware design process, as different stages or configurations may necessitate distinct optimization strategies. Hence, it's crucial to consider these factors when designing hardware to ensure efficient performance across a wide range of scenarios.

Considering these challenges, careful consideration and optimization of hardware designs are necessary. In this section, we will survey and analyze various hardware optimizations  tailored for efficient LLM inference, with a focus on addressing the issues related to varying arithmetic intensity.

\subsection{Spatial Architecture}


The decoding process of LLM involves predicting words one at a time based on previously generated ones. However, this process can be costly, especially during tasks in long sequence generation. This is because the model needs to access large amount of weights and the key-value (KV) cache to generate each token, resulting in low arithmetic intensity.

There are several solutions that have been developed to address this issue. One such solution is the implementation of a "Spatial Architecture". In contrast to traditional computer architectures, spatial architectures utilize a different approach to computing. Instead of folding the computation process into multiple interactions between processing elements (PEs) and main memory, spatial architectures distribute the computation across multiple PEs. This design allows for the exploitation of parallelism, as each PE simultaneously performs a portion of the computation. Additionally, the intermediate data flows between the PEs, avoiding writing back to DRAM each time.

\begin{figure}[t]
    \centering
    \includegraphics[width=0.95\linewidth]{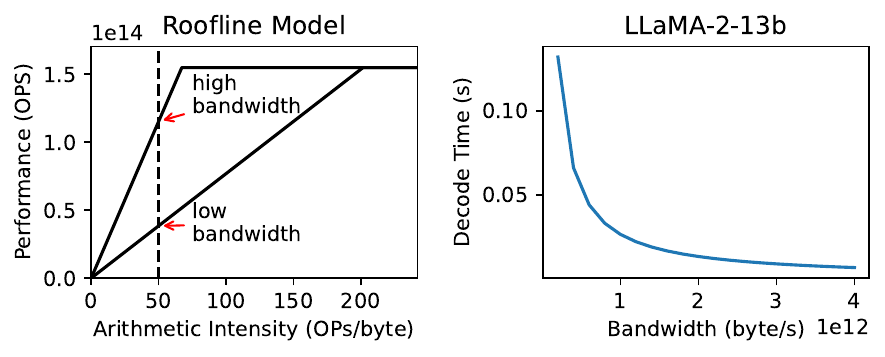}
    \caption{Effect of bandwidth on roofline model and Llama-2-13b. (Batch size=1, sequence length=1024)}
    \label{fig:hardware_bandwidth}
\end{figure}

In a spatial architecture, each PE is responsible for a specific portion of the computation. To facilitate efficient communication, data is typically moved between neighboring PEs. This allows for improved performance and efficient utilization of resources.
In a spatial setup, each PE has its own direct access to memory. This enables multiple processing units to access memory simultaneously, improving the overall speed at which information can move in and out of memory. This results in enhanced memory bandwidth and overall LLM inference performance.
As shown in Figure~\ref{fig:hardware_bandwidth}, as the total memory bandwidth increase, the performance for the linear layer in decoding stage can significantly increase.

In one case, Groq employs their LPU~\citep{abts2022groq_lpu} to create a spatial system for LLM inference. This system achieves a remarkable speed of over 300 tokens per second per user on the Llama-2-70b model~\citep{groq2023lpu_report}. Another example is Graphcore's Intelligence Processing Unit (IPU), which is another type of spatial architecture that efficiently executes LLMs~\citep{graphcore_gradient_huggingface}.



\subsection{Processing in Memory}

The decoding phase of LLM inference experiences the so-called "Memory Wall" problem, primarily due to its low arithmetic intensity. This issue is not new, the computer architecture community has been struggling with the "Memory Wall" problem for several decades. Among various potential solutions, processing in-memory  techniques (PIM) have garnered significant interest in recent years. By placing compute units directly in memory chips, we can leverage the much higher internal memory bandwidth and reduce the data movement overhead between memory and CPU/GPU cores. 

In recent years, DRAM-based PIM has entered the commercialization phase, which can potentially mitigate the memory bandwidth bottleneck of LLM inference. As listed in Table~\ref{tab:pim-compare}, UPMEM's PIM-DIMM~\cite{upmem} is the first commercialized DRAM-PIM product, which places general-purpose RISC cores in DDR4-DIMMs. However, this product was not intended for deep-learning applications, thus the peak bandwidth and throughput can hardly meet the requirements of LLM inference. Compared to UPMEM's PIM-DIMM, Samsung proposes to place MAC units into HBM memory, achieving 2TB/s of internal memory bandwidth, which is much higher than that of traditional HBM2 (307GB/s per cube) memory. Since the processing units are tailored for deep learning workloads, the peak compute throughput of HBM-PIM can reach 1.2TFLOPS. In other words, HBM-PIM is suitable for accelerating operators with 1-2 Ops/Byte of arithmetic intensity. 

 Choi et al has proposed to accelerate the KV-Cache processing using HBM-PIM, which has low arithmetic intensity in batched LLM inference. According to their evaluation~\cite{choi2023unleashing}, A GPU+HBM-PIM system for LLM inference can achieve 3.24$\times$ speedup over traditional  monolithic GPU system.

Similar to Samsung's HBM-PIM, SK-hynix has also proposed a GDDR6-based PIM accelerator~\cite{gddr-pim} called AiM. As shown in Table~\ref{tab:pim-compare}, the compute units of AiM adopts the BF16 data format, which is more efficient for deep-learning acceleration. With optimized MAC units, AiM offers 1TFLOPS per chip of compute capacity, while the peak bandwidth is 1TB/s per chip. Although AiM has not reported its performance on LLM, it can achieve up to 10$\times$ speed improvement compared to GPU+HBM2 system on LSTM tasks. 

Note that, although DRAM-based PIM techniques have demonstrated promising potential of  accelerating memory-intensive operators in LLM inference, there are still some limitations that should be addressed in the future. 

\begin{itemize}
\item \textbf{Limited Computation Power.} A key restriction of DRAM-PIMs in accelerating LLMs  stems from their constrained computational capabilities. DRAM-PIMs utilize compute units crafted using the DRAM process, resulting in transistors that are 3 times slower and logic density several times lower compared to CMOS in the same technology node~\cite{devaux2019true}. Even worse, DRAM chips usually have fewer metal layers, leading to a lower routing density at the same time. Due to these technical constraints, DRAM-PIMs can hardly incorporate powerful compute units. As a consequence, DRAM-PIMs is only suitable for small-batch inference or KV-cache processing. For computation-intensive large-batch inference, a powerful host is still necessary.

\item \textbf{Capacity Constraints.} Another significant limitation of DRAM-PIMs is their restricted capacity. Given that DRAM-PIMs allocate some memory capacity to construct the computation units, the total memory capacity is typically 50\% less than that of standard memory~\cite{hbm-pim}. For LLM applications that necessitate substantial memory capacity to accommodate the weights and KV-cache, DRAM-PIMs may encounter capacity-related challenges.

   \item \textbf{Inadequate Inter-PIM Communication.} In addition to the constraints imposed by computation power and capacity, another limitation of DRAM-PIMs  is their suboptimal inter-PIM communication capability. Given that there are distributed computing units located near each DRAM bank, data aggregation and computation synchronization among these units are unavoidable. However, DRAM-PIMs lack robust interconnects~\cite{zhou2023dimm,jonatan2024scalability}, and they typically depend on the host CPU/GPU for data exchange between PIM units. This reliance can lead to inefficiencies in the system. Therefore, to enhance LLM inference, future iterations of DRAM-PIMs should aim to improve their inter-PIM communication abilities.
\end{itemize}

\input{tables/pim}





\subsection{New Data Format}
\label{sec:dataformat}


Neural networks typically employ high-precision floating-point numbers (16 or 32 bits) for training. While high-precision FP numbers can accommodate both representation precision and range, the complex hardware implementation required for FP arithmetic is not conducive to efficient inference.
To mitigate the hardware overhead, uniform quantization casts high-precision floating-point numbers to low-precision integer representations, substituting expensive floating-point logic with efficient integer logic. However, uniform quantization struggles to balance representation precision and range simultaneously, leading to substantial degradation in model accuracy. 
Moreover, preserving model accuracy without degradation necessitates well-designed quantization algorithms, introducing additional casting efforts.
Non-uniform quantization attempts to enhance the precision of data representation under low-bit conditions by assigning bits and discretizing the range of parameters non-uniformly. Yet a critical drawback of non-uniform quantization is its challenging deployment on general computation hardware, e.g. CPUs and GPUs~\citep{gholami2022survey}.
In summary, existing data formats fail to concurrently achieve fine precision, extensive range, high efficiency, and low tuning costs for inference.
Given the criticality of reducing the deployment costs of LLMs, a substantial amount of work investigate into exploring the most balanced data format tailored to LLMs.

\begin{table}[htbp!]
\centering
\resizebox{\linewidth}{!}{%
\begin{tabular}{cccc}
    \hline
    \textbf{Aspect} & \begin{tabular}{@{}c@{}} \textbf{Floating} \\ \textbf{Point} \end{tabular} & \begin{tabular}{@{}c@{}} \textbf{Uniform} \\ \textbf{Quantization} \end{tabular} & \begin{tabular}{@{}c@{}} \textbf{Non-Uniform} \\ \textbf{Quantization} \end{tabular} \\
    \hline
    Precision        & Good & Bad & Medium \\
    Range            & Good & Bad & Medium  \\
    HW Efficiency    & Bad  & Good & Bad  \\
    Casting Effort    & Good  & Bad & Medium  \\
    \hline
\end{tabular}
}
\caption{Comparison of Floating Point, Uniform Quantization, and Non-Uniform Quantization}
\label{tab:data_format_comparison}
\end{table}

To trade for better hardware efficiency from the original FP model, a natural progression is to reduce the exponent \& mantissa bits in the high-resolution floating-point format.
As demonstrated by recent works~\citep{micikevicius2022fp8, sun2019hybrid}, models of various categories pre-trained in FP16 (including LLMs) can be directly quantized to FP8 without significant accuracy degradation. Moreover, on a wide spectrum of tasks, training with FP8 can effectively match the result quality achieved by 16-bit training sessions.
The significant gain of hardware efficiency and little demand of user efforts envisioned by low-resolution float-point formats have garnered the attention from AI hardware manufacturers.
For instance, NVIDIA implements FP8 Tensor Core in their latest H100 GPUs~\citep{nvidiahopperarchitecture}.
Tesla also introduces configurable floating point format, namely CFloat8, in their Tesla Dojo chip~\citep{tesladojotechnology}.

Beyond the noval architectures introduced by the industry, academia has also initiated efforts to exploiting the potential of low-precision floating-point format on LLMs.
ZeroQuant-FP~\citep{wu2023zeroquant} proposes FP4 and FP8 for weight/activation quantization of LLMs. The authors adopt scaling constraints for weight quantization, achieving efficient weight conversion from FP4 to FP8 and better utilization of the FP8 Tensor Core.
ZeroQuant-(4+2)~\citep{wu2023zeroquant1} and FP6-LLM~\citep{xia2024fp6} proposes weight quantization of LLMs with FP6, and provides efficient implementation on CUDA Core and Tensor Core, respectively. 
LLM-FP4~\citep{liu2023llm} proposes quantizing both weights and activations of LLMs down to FP4.
In summary, these efforts demonstrates the feasibility of applying floating-point formats at even lower bit-widths for quantization, as well as the potential for achieving greater efficiency gains on existing or new hardware platforms.

On the other hand, researchers are delving into the refinement of low-precision quantization formats to augment the adaptability of data representation whilst preserving hardware efficiency. 
One line of works propose to explore new encoding schemes within single value representation.
Contrary to INT and FP numbers, which utilize fixed-length sub-fields for encoding distinct pieces of information, such as exponents and mantissas, the newly-proposed rule-based quantization formats enable dynamic adjustment of sub-field bit-widths. 
ALPS~\citep{langroudi2021alps} proposes a generalized posit format along with a new adaptive quantization algorithm to optimally represent the dynamic range and distribution of DNN parameters.
ANT~\citep{guo2022ant} proposes a new data format called flint with leading-1 encoding for exponent fields.
Dybit~\citep{zhou2023dybit} proposes to separate exponent and mantissa fields with the first encountered 0 as delimiter.
The flexibility of these variable-length data formats presents the chance to compromise between range and precision more effectively and allows for customization to more closely align with the distribution of LLMs' weights and activations.

Another line of works exploit similarity and discrepancy across values.
Outlier-aware quantization capitalizes on the observation that values with large magnitudes have a significant impact on model performance. In this approach, important values are identified as outliers and are treated differently from normal values to ensure a more accurate representation. 
OLAccel~\citep{park2018energy} and GOBO~\citep{zadeh2020gobo} separately store and allocate higher bit-width to the outlier values.
OliVe~\citep{guo2023olive} refines the concept with outlier-victim pair encoding scheme to ensure aligned memory access and improved efficiency.
Bit-sharing encoding concentrates on the inherent similarity among values and annotates additional information in coarse granularity, thereby achieving a balance between representation accuracy and hardware efficiency. 
AdaptivFloat~\citep{tambe2020algorithm} proposes to optimally shift the available range of FP values with a common tensor-wise exponent bias.
MX~\citep{darvish2023shared} extends AdaptivFloat's observation into finer granularity, and proposes Block Data Representation (BDR) framework to explore the optimal tradeoff between representation accuracy and hardware efficiency.

\begin{table}[htbp!]
\centering
\resizebox{\linewidth}{!}{%
\begin{tabular}{ccc}
    \hline
    \textbf{Method} & \textbf{Description} & \textbf{Benefits} \\
    \hline
    Low-bit floating-point & Reduce exponent \& mantissa bits & Efficiency $\uparrow$ \\ 
    Variable-length encoding & Dynamically adjust sub-field bit-width & Precision $\uparrow$ \\
    Outlier-aware quantization & Customize quantization for outliers & Precision $\uparrow$ \\ 
    Bit-sharing encoding & Share common information within a block & Precision $\uparrow$ \\ 
    \hline
\end{tabular}
}
\caption{Summary of improvements in data formats. The benefits comes with negligible negative impacts on other factors.}
\end{table}

\subsection{New Processing Element}


Except for the high demand for memory access, there has been a growing interest in developing specialized processing elements (PEs) to boost the computation. 
These specialized architectures aim to provide significant computational enhancements over general-purpose processing element, such as CUDA core, for the specific operations associated with LLMs.

NVIDIA has developed a special hardware acceleration engine called the Transformer Engine in their H100 GPU. This engine uses statistical analysis to determine the optimal precision (FP16 or FP8) for each layer of the model, achieving the best performance while maintaining accuracy.
Some researchers have designed accelerators specifically for efficiently executing the attention mechanism in language models (LLMs)~\citep{kao2023flat,qin2023fact}.
Several companies and research groups have been exploring the use of FPGAs to speed up LLM computations. Examples include DFX~\citep{hong2022dfx} and LightLLM~\citep{zeng2024flightllm}.


%% file: tables/pim.tex
\begin{table}[t]
\centering
\caption{Comparison of Commodity DRAM NMC Products}
\label{tab:pim-compare}
\resizebox{0.475\textwidth}{!}{
\begin{tabular}{|c|c|c|c|}
\hline
Product         &  PIM-DIMM & HBM-PIM          & AiM            \\ \hline
Technique  &       DDR4         & HBM2   & GDDR6  \\ \hline
PIM Units &  RISC Cores   & FP16 MAC   &  BF16 MAC      \\ \hline
Peak Bandwidth  &   80.4 GB/s    per DIMM          & 2 TB/s per cube   & 1 TB/s per chip  \\ \hline
Peak Throughput &   43.8 GOP/s per DIMM             & 1.2 TFLOPS & 1 TFLOPS \\ \hline
\end{tabular}
}
\end{table}

%% file: sections/8_discussion.tex
\section{Discussion}

\subsection{Reliability}

The discussion above significantly enhances the inference and training efficiency of LLMs in practical scenarios. However, these compression methods also lead to subtle changes in model reliability. Overall, the various compression techniques discussed in Section~\ref{sec:compression} will have a significant impact on model reliability. Therefore, this section primarily focuses on how key design choices within these different compression techniques affect  the following three aspects of reliability: hallucination, safety alignment, and out-of-distribution generalization.

Hallucination primarily refers to cases where the outputs of LLMs do not align with real-world knowledge, often generating content that is either factually incorrect or nonsensical \cite{huang2023survey}. Safety alignment focuses on the model's inherent ability to autonomously recognize and refuse to respond to harmful inquiries, thereby safeguarding against generating inappropriate or dangerous content \cite{ouyang2022training}. Reliability pertains to the model's stability when confronted with unconventional data in long-tail scenarios, such as interference from adversarial examples or with decision shortcuts \cite{geirhos2020shortcut}. In the following parts, we will delve into how different compression methods impact these three crucial aspects of model performance.

\subsubsection{Hallucination}
The ability of an LLM to suppress hallucinations is critically affected by modifications to its parameters. According to previous research findings, factual knowledge is often stored within the Feed-Forward Networks (FFNs) of Transformer modules. Therefore, when employing quantization or structured compression methods, special attention should be paid to the output calibration of the FFN layers. A viable approach to addressing this concern involves identification of key FFN layers, i.e., utilizing neuron-level interpretation techniques from prior research \cite{DBLP:conf/nips/MengBAB22} to identify the FFN layers that are crucial for storing knowledge. These layers are deemed important because they contain weights and representations that are integral to the model's ability to recall and utilize factual information accurately. 

For parts identified as crucial in storing knowledge, the quantization precision should be selectively enhanced. This means that while the overall model undergoes quantization to reduce its size and computational requirements, the quantization process for these critical FFN layers is adjusted to maintain a higher level of precision. This selective approach helps in preserving the integrity and accuracy of the stored factual knowledge, thereby mitigating the risk of output hallucinations.

In the context of pruning, which involves removing weights or neurons deemed less important to streamline the model, it is essential to retain the identified important FFN layers. By preserving these layers, the model maintains its core capability to recall and process factual knowledge, which is essential for ensuring the accuracy of its outputs and reducing the likelihood of generating hallucinated content.

\subsection{Safety Alignment}
Based on previous research findings \cite{yuan2023rptq}, moderate model compression, such as 8-bit quantization, does not significantly compromise the safety capabilities of models. However, it may render models more susceptible to certain jailbreak attacks—a direction that prior studies have seldom covered \cite{deng2023jailbreaker}. Consequently, we recommend conducting comprehensive red teaming before deploying these compressed models. Moreover, knowledge transfer-based approaches can substantially weaken the safety of models. Therefore, we advise re-finetune smaller models after completing knowledge transfer.

\subsection{OOD Generalization}
Large language models, when deployed in real-world scenarios, are often influenced by decision shortcuts, leading to erroneous decisions within the long-tail subgroup distributions \cite{geirhos2020shortcut}. As demonstrated by previous research \cite{yuan2023benchmarking}, neural networks subjected to quantization compression exhibit significant performance disparities across different subgroups within the same task, with errors in judgment frequently occurring in long-tail subgroups that rely on decision shortcuts present in the context. Furthermore, kv-cache compression, a commonly used technique to enhance the inference efficiency of large language models in practice, relies on randomly discarding tokens in the attention matrix during inference. This method further exacerbates the model's reliance on decision shortcuts. Therefore, it is advisable to consider integrating corresponding robustness enhancement methods in downstream specific scenarios, such as the invariant test-time optimization techniques mentioned in prior studies \cite{ma2024invariant}.
\subsection{Efficient Large Multimodal Models}

\subsubsection{Large Multimodal Models (LMMs)}
Large Multimodal Models (LMMs), particularly Visual Language Models (VLMs), have emerged as a promising avenue for creating general-purpose assistants, showcasing significant enhancements in perception and reasoning capabilities. These models leverage LLMs as their cognitive core, enriching multimodal (MM) tasks with robust language generation, zero-shot transfer capabilities, and In-Context Learning. Foundation models across different modalities provide high-quality representations. A critical challenge for LMMs is integrating LLMs with models from other modalities to facilitate collaborative inference effectively. The primary focus has been on improving modality alignment and aligning with human intentions through a MM Pre-Training + MM Instruction-Tuning pipeline. Two surveys, \citep{yin2023survey,zhang2024mm}, delve into LMMs in detail.

\subsubsection{Efficient LMMs}

The need for cross-modality capabilities in resource-limited scenarios has become increasingly apparent. Despite LMMs' advancements, their large-scale training and deployment incur significant computational costs, necessitating efficient parallel device implementations. Google's Gemini~\citep{team2023gemini} leads in efficient LMMs, achieving state-of-the-art performance on multimodal benchmarks and introducing mobile-scale LMMs suitable for low-memory devices. However, Gemini remains closed-source. Open-source initiatives, like LLaVA-v1.5, utilize advanced compression techniques, such as 4/8 bit quantization via bitsandbytes~\citep{dettmers2022llm}, for more on compression techniques, see Section~\ref{sec:compression}.

Further efforts towards efficient LMMs include MobileVLM~\citep{chu2023mobilevlm}, which develops compact LLMs and an efficient multimodal feature projector, and its successor, MobileVLM-v2~\citep{chu2024mobilevlm}, which explores improved training strategies for mobile scenarios. TinyGPT-V~\citep{yuan2023tinygpt} leverages the advanced Phi-2~\citep{javaheripi2023phi} LLM to surpass the performance of significantly larger models. 
Similarly, LLaVA-Phi~\citep{zhu2024llava} and Vary-toy~\citep{wei2024small} introduce smaller backbones and enhanced vocabularies for broader generalizability. 
TinyLLaVA~\citep{zhou2024tinyllava} investigates the impacts of architectural choices, data quality, and training strategies, demonstrating that smaller LMMs can match the performance of their larger counterparts with optimized data and training. MoE-LLaVA~\citep{lin2024moe} adapts Mixture of Experts (MoE)~\citep{6215056} to mitigate model degradation due to sparsity. 

\subsection{Long Context Modeling} 
When used for tasks like Chatbot or document summarization tools, Large Language Models' long context language modeling and reasoning capabilities are challenged. The model, however, are usually trained on the general pretraining corpus that usually comprises text snippets, not long enough to serve as high-quality training examples for LLM to learn. To alleviate the insufficient long context capabilities of the pre-trained models, abundant works have attempted to attach the problem from different angles. For this section of the discussion, we mainly focus on alternative attention mechanisms in \ref{efficient_attention}, cache compression and context retrieval, and position encoding modifications. For more detailed studies and a more comprehensive review of the issue of LLM long-context modeling, we refer interested readers to the recent survey specifically on this topic \cite{huang2023advancing}. 

\subsubsection{Alternative Attention Design} \label{efficient_attention} 
Lying in the core of transformer architecture is the self-attention mechanism. For decoder-only model inference with KV cache, if the past context is long, attending to all past keys computing with past values incurs both computation and memory bottleneck. Prior works found that not all past tokens need to be attended to preserve the model inference performance. Landmark Attention (\cite{mohtashami2023landmark}) introduced the special landmark attention into the sequence to summarize the following block of the tokens' information. The new query would attend to the landmark tokens first to determine whether the following tokens inside the block are needed for predicting the next word, thus reducing the attention computation while maintaining the random-access nature of attention. 
Earlier, Funnel-Transformer (\cite{dai2020funnel}) also approaches the same goal of conducting attention in the block level. Differently, they introduce a down-sampling during the encoder portion and an up-sampling method during the decoder portion. The reduced FLOPs allow them to build a deeper and wider model that outperforms the original model with the same compute budget. 
On the other hand, Longformer (\cite{beltagy2020longformer}) gives an early attempt to combine sliding window attention and global attention, in which they only allow a few pre-defined tokens to attend to all tokens in the sequence, while other tokens do sliding window attention and also attend to these selected global attending tokens. Concurrently, ETC (\cite{ainslie2020etc}) introduces a 2-level hierarchy to the input so that the plain long input tokens can do sliding window attention, while a global input of auxiliary tokens extracted and downsampled from the original input can do normal attention. The plain inputs are allowed to attend to the global inputs, thus acquiring the global context information. Similarly, LongT5 (\cite{guo2022longt5}) proposes an even simpler method to directly downsampled the long and prior context using mean pooling with chunksize of 16, the downsampled keys and values are directly appended in front of the rest of the input sequence. Then the rest of the model can do sliding window attention with the additional attention to the downsampled context summary tokens at the front. 
Under the LLM era when pretraining becomes prohibitively costly, StreamingLLM (\cite{xiao2023efficient}) and LM Infinite (\cite{han2023lm}) concurrently propose a sink plus sliding window attention pattern to LLM as a plug-in tool to boost the LLM's long context ability. StreamingLLM in particular points out that due to the mechanism of softmax operation in the transformer, the beginning tokens of the input are crucial to maintaining the self-attention performance. The modified attention mask is shown to achieve strong results in long-context language modeling without additional model finetuning. Besides, $H_{2}O$ (\cite{zhang2024h2o}) reduces the self-attention computational complexity by only attending to tokens of interest out from the prior prefilled input context. To do that, they build an empirical oracle for selecting tokens. The method is shown to benefit LLM in long-context regime. 

\subsubsection{Recurrence and Retrieval} 
Transformer-XL \cite{dai2019transformer} proposes to introduce the segment-level recurrence structure to the Language Model to boost the current language model in long-context capabilities. The method stores the last layer output of one previous segment to append to the current layer, thus drastically increasing the dependency distance of the model. Segatron (\cite{bai2021segatron}) and Compressive Transformer (\cite{rae2019compressive}) extend upon the previous idea. Segatron boosts the segment-level recurrence from the position embeddings on the token level, sentence level, and beyond through segment-aware mechanisms. Compressive Transformer proposes a second-level compressed memory FIFO queue so that the past segment context should not be discarded but compressed by their customized function and stored in the queue to elongate the length of context dependency. 
In the LLM era, Dynamic Memory Compression \cite{nawrot2024dynamic} also follows the recurrence idea with compressed context to dynamically decide how to compress the previous context information, thus reducing the sequence length of the attention while preserving the distant information. Besides, other than segment-level, \cite{fan2020addressing} studies the retrospective recurrence. Memorizing Transformer (\cite{wu2022memorizing}) and Memformer (\cite{wu2020memformer}) combines retrieval and local cache with a forgetting mechanism.  \\ 

\noindent On the other hand, following on trend that not every past token is needed for the current token generation next, the past KV cache can be placed physically further in distance, i.e. secondary memory, so that when only needed, the specific key-value pair will be retrieved. Thus, another way to boost the LLM is through Retrieval-Augmented Generation (RAG). RAG is a heated topic in its own light and contains abundant techniques and past works. Due to the limited scope of our paper, we kindly refer to the following comprehensive surveys on RAG for interested readers: \cite{gao2023retrieval} and \cite{zhao2024retrieval}. To address the particular motivation underlined long-context capabilities brought up at the beginning of this section, LangChain \cite{pandya2023automating} is the popular way to mitigate the Chatbot's past long conversation through retrieval. LangChain is an open-source tool that specializes in using LLM to compute embedding for user-input long documents and files, so that later according to the user's prompt, top-relevant contents will be retrieved through the cosine-similarity metrics. Besides, there are fast-growing other relevant works(\cite{borgeaud2021improving}, \cite{bertsch2024unlimiformer}, \cite{zhong2023memorybank}, \cite{zhou2023recurrentgpt}, \cite{kynoch2023recallm}, \cite{modarressi2023retllm}, \cite{guu2020retrieval}, \cite{wang2024augmenting}) under the retrieval for long-context settings. 

\subsubsection{Maneuvering Position Encodings} 
During pretraining, the position encoding of the transformer hasn't seen an input sequence length longer than a fixed limit. Also, because the position encoding is usually based on a triangular function, the vanilla transformer is unable to extrapolate to unfamiliar longer sequence lengths. Earlier techniques add attention bias to the attention map before the softmax operation. ALiBi (\cite{press2021train}) introduces heuristics to design such attention bias, achieving early success in long-context extrapolation tasks for transformer architecture. Besides, Kerple (\cite{chi2022kerple}) and Sandwich building on the prior works, introduce trainable parameters to construct the attention bias matrix or construct attention bias through sinusoidal properties of position encoding. \\ 

\noindent On the other hand, another lively line of work delves into adjusting the RoPE. Inspired by the Neural Tangent Kernel (NTK) theory, NTK-aware Scaled RoPE (\cite{xiong2023effective}) modifies the base parameter of the RoPE; LEX (\cite{sun2022lengthextrapolatable}) and  PermuteFormer (\cite{chen2021permuteformer}) adds an exponential decay term; Positional Interpolation places linear scaling to every token; Dynamic-NTK (\cite{huang2019dynamics}) gradually increases the scaling ratio. In the LLM era, YaRN (\cite{peng2023yarn}) linearly scales query and key with a temperature factor. Giraffe (\cite{pal2023giraffe}) finds that high-frequency terms are hurt by the low-frequency terms that are usually ill-trained and proposes a scaling mechanism based on the power law to protect the well-trained high-frequency information. 


%% file: sections/9_conclusion.tex
\section{Conclusion}
In this work, we review on efficient large language model (LLM) inference. 
For this practice-driven topic, our comprehensive study goes beyond conventional literature reviews by providing both an overview of existing research and the development of a roofline model. 
Our first step is to develop a roofline model, which enables us to pinpoint bottlenecks in LLM deployments, allowing researchers to resort to more specific deployment strategies. 
By meticulously assembling the latest developments in the field, our survey spans an array of pivotal areas, including innovations in weight optimization techniques, enhancements in decoding algorithms, as well as advancements in hardware and system-level optimizations.
It is important to note that this project will be updated and maintained.